%% file: main.tex
\documentclass[11pt]{article}

\usepackage[final]{acl}

\usepackage{times}
\usepackage{latexsym}

\usepackage[T1]{fontenc}

\usepackage[utf8]{inputenc}

\usepackage{microtype}

\usepackage{inconsolata}

\usepackage{graphicx}
\usepackage[table]{xcolor}
\usepackage{geometry}

\usepackage{balance}      %
\usepackage{booktabs}     %
\usepackage{caption}      %
\usepackage{array}
\usepackage{enumitem}
\usepackage{multirow}
\usepackage{subcaption} %

\usepackage{algorithm}        %
\usepackage[noend]{algpseudocode} %

\usepackage{float}

\usepackage{amssymb,amsfonts}%
\usepackage{amsmath}

\title{Faithful Persona Steering under Incongruity via Dual-Stream Refinement}

\author{Yu-An Chu \and Jen-Ren Pong \and Chia-Yao Yeh \and Meng-Fen Chiang\textsuperscript{\textbf{*}}\\
  National Yang Ming Chiao Tung University \\
  Hsinchu, Taiwan \\
  \texttt{\{chu1004.ee11, jenrenpong.ee14, jiayao920406.ee14, meng.chiang\}@nycu.edu.tw} \\}

\newcommand\blfootnote[1]{%
  \begingroup
  \renewcommand\thefootnote{}\footnote{#1}%
  \addtocounter{footnote}{-1}%
  \endgroup
}  

\begin{document}
\maketitle

\blfootnote{$^*$Corresponding author.}

\begin{abstract}
Standard LLM personalization typically frames identity as a static retrieval task, overlooking the inherent \textit{incongruity} of human personas, where stable traits coexist with atypical, context-specific stances. Existing methods struggle to reconcile these dimensions: prompting succumbs to \textit{context drift} over long sequences, while fine-tuning often suppresses idiosyncratic ``quirks'' in favor of generic distributional patterns. To bridge this gap, we present \textsc{QuirkyMind}, a framework that disentangles identity \textit{definition} from its \textit{expression}. First, \textit{Traits Anchoring} constructs a dual-stream latent state, fusing a sentence-level summary for semantic stability with a token-level sequence for generative control. This state is stabilized via \textit{In-Context Narrative Refinement} using an alternating objective: a discriminative InfoNCE loss anchors the persona in representation space to prevent drift, while a generative cross-entropy loss ensures faithful verbalization. Finally, \textit{Persona Steered Generalization} transfers the refined state to downstream tasks via parameter-efficient adapters. Empirical evaluations on Persona-Steered QA and Narrative Inference demonstrate that QuirkyMind mitigates drift, consolidating persona knowledge without erasing authentic incongruities.
\end{abstract}

\input{ACL-main/01-intro}
\input{ACL-main/02-preliminary}

\input{ACL-main/03-method}

\input{ACL-main/04-experiment}
\input{two-column/05-related-work}

\section{Conclusion}
We presented \textsc{QuirkyMind}, a framework addressing persona incongruity by disentangling semantic identity from generative expression. By anchoring traits in a dual-stream latent state refined via alternating objectives, the model resists context drift while preserving the ``quirks'' essential for authentic personas. Empirical evaluations on PQA and PNI demonstrate that treating incongruity as a structural signal ensures both consistency and narrative stability, establishing a foundation for resilient, sociologically grounded personalization.

\section*{Acknowledgments}
This work is supported by the National Science and Technology Council (NSTC), Taiwan (114-2222-E-A49-004 and 114-2639-E-A49-001-ASP).

\section*{Limitations}
While QuirkyMind robustly handles persona incongruity, several limitations remain.

\noindent \textbf{Geocultural and Sociodemographic Scope.} The empirical scope of this work is circumscribed by its reliance on the American Trends Panel (ATP) and PersonaChat, both of which predominantly reflect Western ideological and sociodemographic paradigms. Consequently, the framework’s efficacy within non-Western cultural contexts or divergent identity systems remains to be established. Future research should prioritize developing cross-cultural benchmarks to ensure broader sociopolitical generalizability.

\noindent \textbf{Representational Capacity.} Although our dual-stream approach mitigates context drift, extreme or highly divergent dialogue trajectories may still encounter representational bottlenecks. Specifically, a single latent state $\textbf{z}^{(\star)}$ may prove insufficient to capture the nuance of highly multifaceted or internally contradictory identities during prolonged interactions, where the required persona depth exceeds the current state's capacity.

\noindent \textbf{Static vs. Dynamic Persona Alignment.} This study focuses exclusively on alignment with static backstories. We have yet to investigate how refined persona states evolve or adapt dynamically in long-form, multi-agent strategic dialogues, where identities may shift in response to adversarial pressure or collaborative negotiation.

\bibliography{ACL-main/ref}
\appendix
\newpage
\input{two-column/07-appendix}

\end{document}

%% file: ACL-main/01-intro.tex
\section{Introduction}\label{sec:intro}

\begin{figure}[t]
    \centering
    \includegraphics[width=1\linewidth]{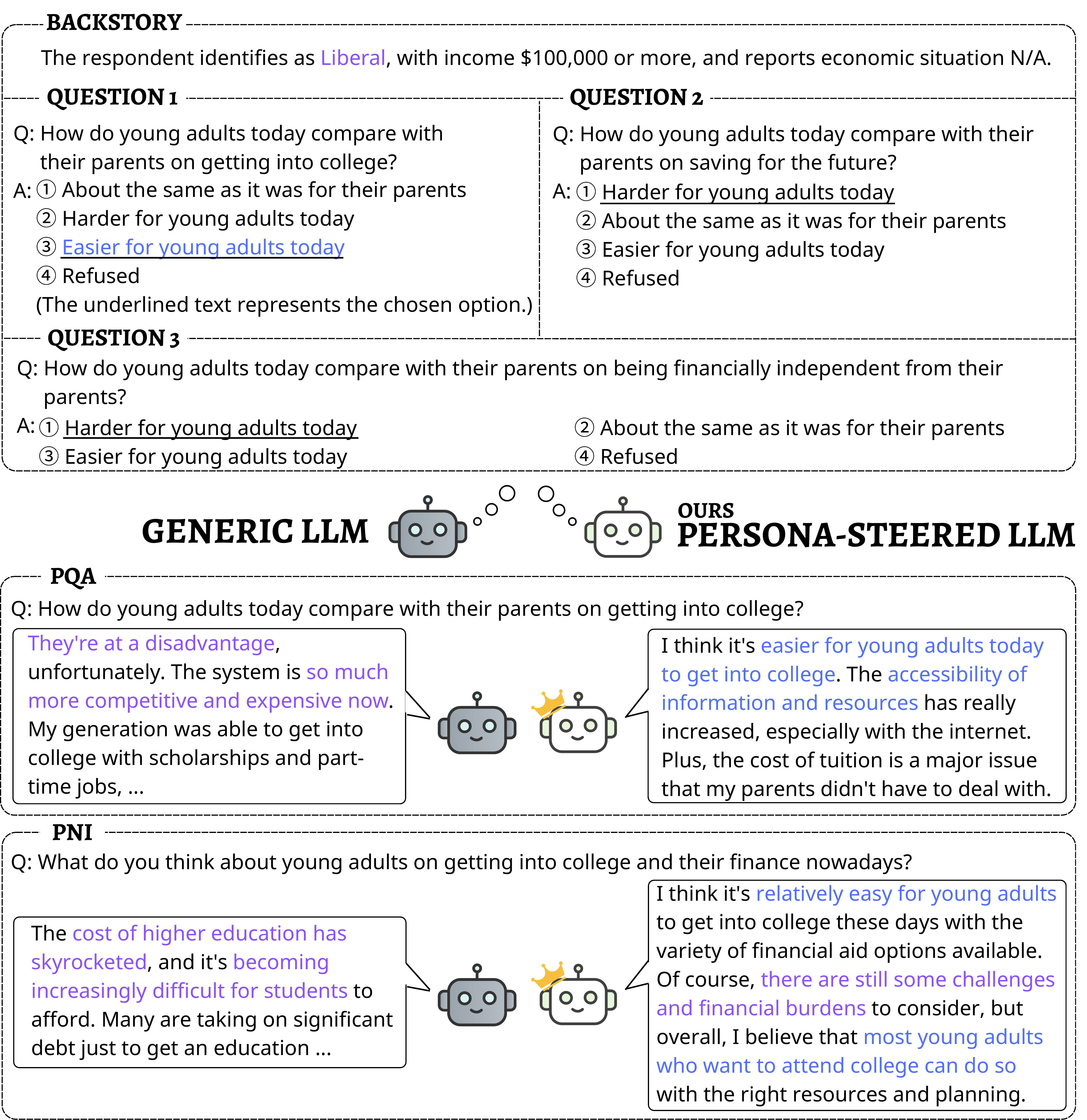}
    \caption{\textbf{Faithful Persona Steering under Incongruity.} 
(a) \textbf{Baselines (Llama 3.1-8B):} Standard prompting methods succumb to \textit{context override} ; salient but atypical stances (incongruities) fade over long contexts, causing the model to drift toward generic priors.
(b) \textsc{QuirkyMind (Ours):} Anchors traits via a stable latent state refined by in-context evidence. This mitigates drift, preserving consistent performance in PQA and PNI even when identities defy stereotypes.}
    \label{fig:motivation}
\end{figure}

Personalization in Large Language Models (LLMs) is frequently treated as a static conditioning task. This view overlooks the reality that identity is not a fixed enumeration of attributes but a dynamic construct that unfolds through dialogue. A speaker can be simultaneously consistent with a long-term backstory yet surprising in specific contexts . A critical failure mode of current methods is their inability to model \textit{persona incongruity}, namely, instances where stable values collide with atypical stances (e.g., a ``climate activist'' who supports ``nuclear energy''). Figure~\ref{fig:motivation} illustrates how baselines drift into generic responses under such tension, whereas our framework maintains faithful steering.

Existing methods struggle to reconcile these tensions. Approaches relying on prompt engineering or Retrieval-Augmented Generation (RAG) often succumb to \textit{context drift}, where the persona fades over long contexts, or \textit{context override}, where the model favors recent tokens over the foundational backstory. Conversely, fine-tuning approaches tend to smooth the distribution, averaging out idiosyncratic traits to fit generalized patterns \citep{choi2024picle}. Consequently, models fail to control where the persona resides in the representation space, leading to inconsistent answers or loss of distinctiveness.

To tackle the challenge of faithful steering, we propose \textsc{QuirkyMind}, a novel framework for in-context persona learning that robustly models both congruous and incongruous traits. Our key insight is to disentangle the \textit{definition} of identity from its \textit{expression}. QuirkyMind operates via three stages. First, \textbf{Traits Anchoring} constructs a robust initial persona state by fusing a sentence-level summary (semantic stability) with a token-level sequence (generative control) into a single latent state. Second, \textbf{In-Context Narrative Refinement} stabilizes this state through simulated multi-turn dialogue. We employ a dual-objective strategy: a discriminative InfoNCE loss anchors the persona in representation space to prevent drift, while a generative Cross-Entropy loss ensures faithful verbalization in token space. Finally, \textbf{Persona Steered Generalization} transfers this refined state to downstream tasks using parameter-efficient lightweight adapters, keeping the backbone frozen.
Empirical evaluations across single and multi-turn settings demonstrate that QuirkyMind significantly outperforms baselines in Persona-Steered Question Answering (PQA) and Narrative Inference (PNI). It successfully consolidates persona knowledge without erasing necessary incongruities, maintaining alignment with both the backstory and issue-specific stances.
Our contributions are as follows:
\begin{itemize}[leftmargin=1em]
    \item We formalize the challenge of \textit{persona incongruity}, revealing how existing methods fail to reconcile stable traits with atypical stances.
    \item We propose \textsc{QuirkyMind}, a framework coupling semantic stability with expressive control via an alternating discriminative-generative objective.
    \item Evaluations across diverse backstories show that our approach yields superior alignment and narrative coherence without suppressing the ``quirks'' that render a persona authentic.
\end{itemize}

%% file: ACL-main/02-preliminary.tex
\section{Preliminaries}\label{sec:pre}

\subsection{Background}

\paragraph{Persona Definition.}
A \textit{Persona} $P$ is a textual specification conditioning a language model toward the characteristics of a specific virtual agent. We instantiate a persona via a \textit{backstory} $\mathcal{B}$, a first-person narrative encoding attributes such as demographics, values, and beliefs. $\mathcal{B}$ serves as the foundational context to steer the model $M$.

Let $Q$ be a set of dialogue questions, and let $Q_{\mathcal{B}} \subseteq Q$ be the subset of questions relevant to backstory $\mathcal{B}$. For a question $q \in Q_{\mathcal{B}}$ with a finite answer set $\mathcal{A}_q$, let $Y_q \in \mathcal{A}_q$ denote the human response. We define the empirical distribution of responses given the backstory as:
\begin{equation}
\label{eq:response-dist}
    p_{\mathcal{B},q}(r) = \Pr(Y_q = r \mid \mathcal{B}, q), \quad r \in \mathcal{A}_q,
\end{equation}
where $r$ ranges over the finite answer set $\mathcal{A}_q$, and $Y_q$ denotes the human response to question $q$ given backstory $\mathcal{B}$

\paragraph{Congruity and Incongruity.}
We formally distinguish between expected and unexpected persona behaviors.From a causal perspective, persona incongruity can be viewed as a counterfactual intervention on latent belief variables \citep{pearl2009causality}. Given a backstory $\mathcal{B}$, \textit{Persona Congruity} over a question set $Q_{\mathcal{B}}$ is the rate at which the model aligns with high-probability human responses:
\begin{equation}
\label{eq:congruity}
    \text{Congruity}(\mathcal{B}) = \frac{1}{|Q_{\mathcal{B}}|} \sum_{q \in Q_{\mathcal{B}}} \mathbb{1}(\hat{Y}_q \in \mathcal{C}_q),
\end{equation}
where $\mathcal{C}_q \subseteq \mathcal{A}_q$ denotes the set of high-probability (congruous) responses for question $q$, and $\hat{Y}_q$ is the model's predicted answer.
Conversely, \textit{Persona Incongruity} refers to instances where valid stances defy the model's learned priors. A response is characterized as \textit{incongruous} if the ground truth $Y_q$ occupies a low-likelihood region of the prior empirical distribution yet remains the true stance of the persona.

\begin{figure}[t]
\centering
\includegraphics[width=1\linewidth]{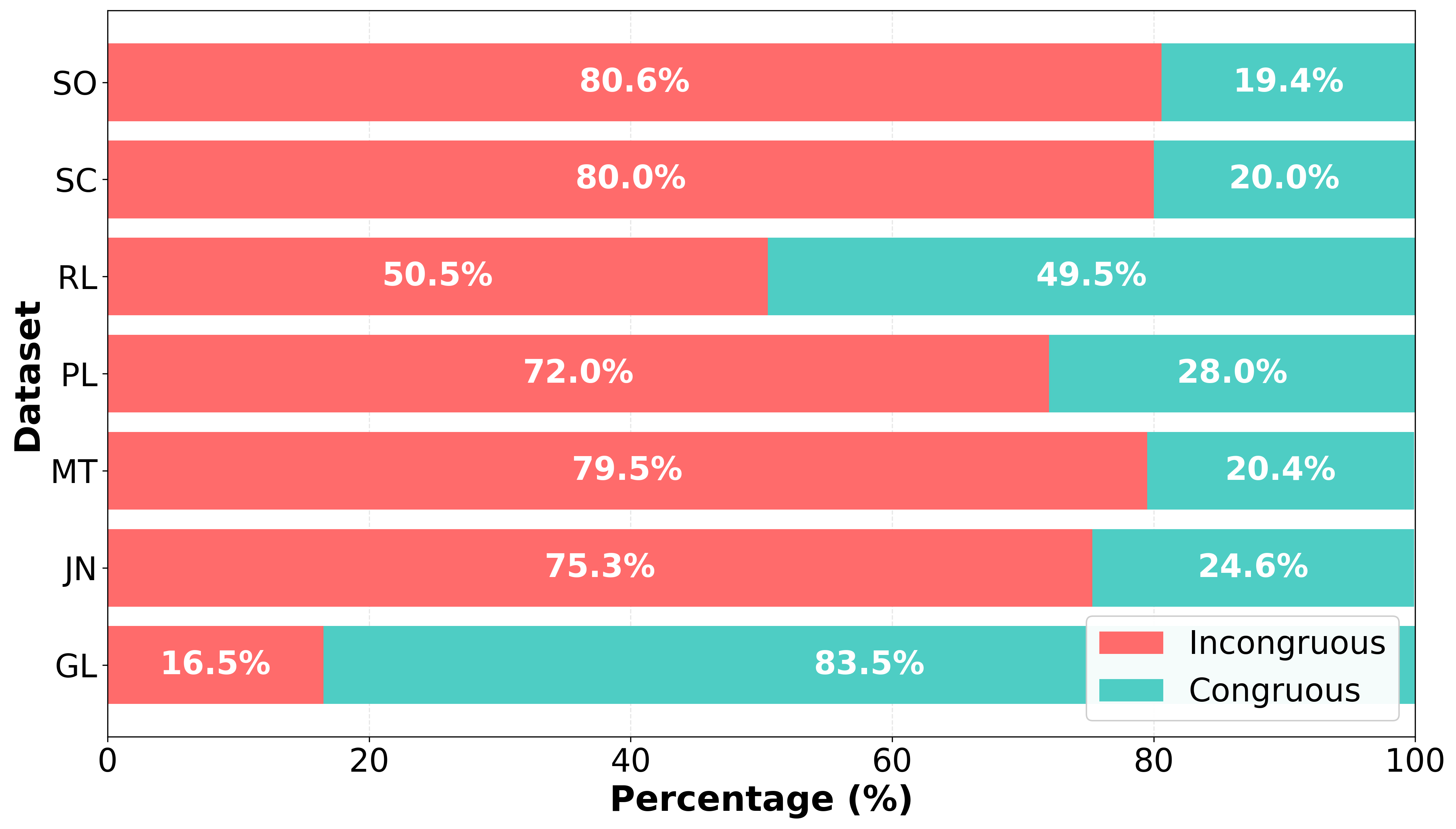}
\caption{\textbf{Distribution of Persona Congruence.} We map the prevalence of congruous versus incongruous personas across seven topical domains in the American Trends Panel (ATP). The data reflects sociodemographic complexity, where authentic stances often diverge from stereotypical group attributes.}
\label{fig:domain_distribution}
\end{figure}

\begin{figure*}[!htbp]
    \centering
    \includegraphics[width=1\linewidth]{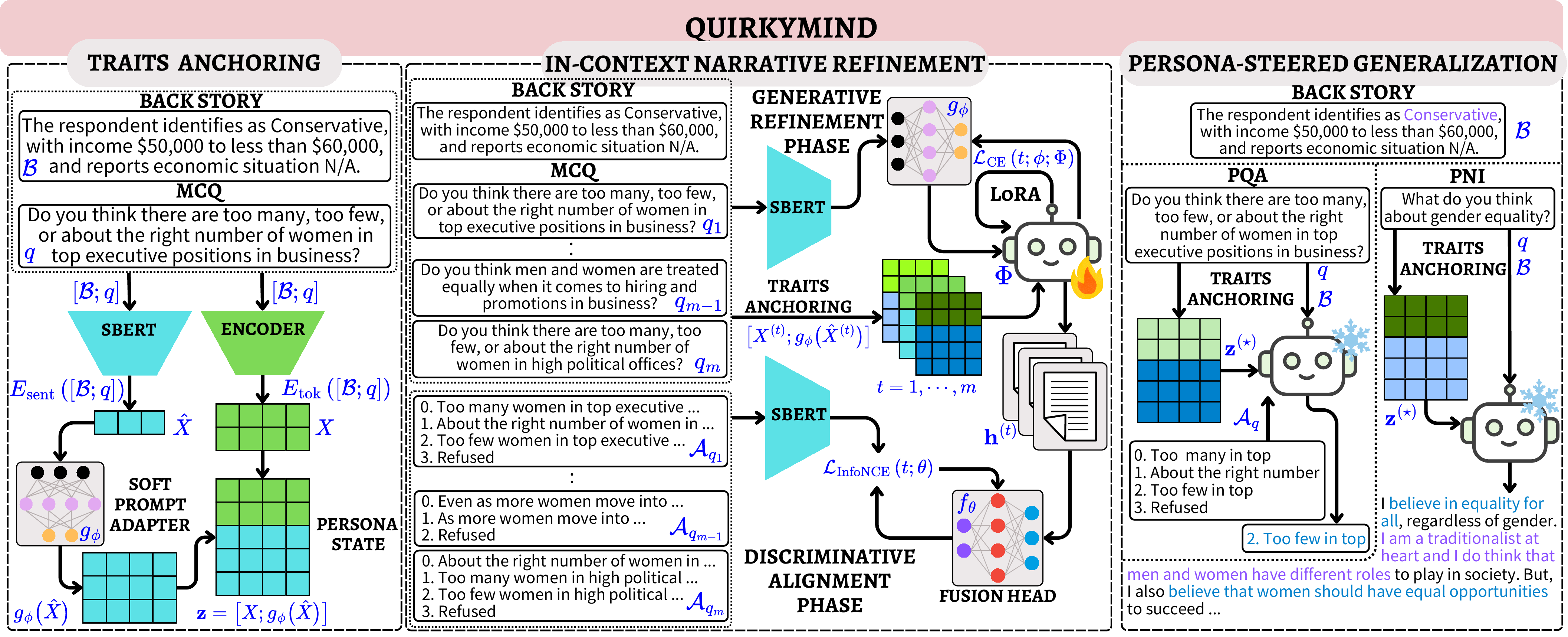}
    \caption{Architecture Overview of \textsc{QuirkyMind} Framework: In-context Persona Learning and Transfer. 
    }
    \label{fig:framework}
\end{figure*}

\noindent \textbf{Example:}
Consider $\mathcal{B}$: ``I am a climate-focused activist who supports strong emissions policy.'' For a question $q$ on phasing out coal, the empirical majority typically favors ``Yes.'' However, if the specific persona prioritizes grid stability, the ground truth might be ``No.'' Here, $p_{\mathcal{B},q}(\text{No})$ is low, rendering the stance incongruous but accurate. Standard backbones frequently succumb to mode collapse here, defaulting to the generic, congruous response. Figure~\ref{fig:domain_distribution} details these types. Crucially, significant incongruity exists across both high- and low-resource domains, challenging models to generalize beyond simple pattern matching.

\subsection{Problem Formulation}

\paragraph{Persona Representation Learning.}
Let $\mathcal{D} = \{(\mathcal{B}, q, r)\}$ be a corpus of triples containing a backstory, a question, and a target response. Our objective is to learn a parameterized encoder $\Gamma_\theta$ that maps the backstory and interaction history to a latent persona state $\mathbf{z}$. This state conditions a language model $M_\Phi$ to approximate the target response distribution. We evaluate the quality of $\mathbf{z}$ on two downstream tasks:

\paragraph{Persona-Steered Question Answering (PQA).}
For $q \in Q_{\mathcal{B}}$ with answer options $\mathcal{A}_q$, the model defines a distribution $P_\Phi(r \mid q, \mathcal{B}, \mathbf{z})$. The task is to predict the categorical response $\hat{Y}_q$:
\begin{equation}
\label{eq:pqa}
    \hat{Y}_q = \operatorname{argmax}_{r \in \mathcal{A}_q} P_\Phi(r \mid q, \mathcal{B}, \mathbf{z}),
\end{equation}
where $P_\Phi$ is the conditional distribution of the LLM parameterized by $\Phi$, and $\mathbf{z}$ is the latent persona state. Our objective is to optimize $\Phi$ to maximize predictive accuracy, ensuring the model remains faithful to the persona’s specific stances even when they deviate from the base model's priors.

\paragraph{Persona-Steered Narrative Inference (PNI).}
Given $\mathcal{B}$, $\mathbf{z}$, and an open-ended prompt $q$, the model generates a free-text response $\mathbf{x}$ sampled via:
\begin{equation}
\label{eq:pni}
    P_\Phi(\mathbf{x} \mid q, \mathcal{B}, \mathbf{z}),
\end{equation}
where $\mathbf{x} = (x_1, \ldots, x_N)$ is the generated token sequence, $q$ an open-ended prompt, and $\mathbf{z}$ the persona state. The model is optimized so that $\hat{\mathbf{x}}_q$ faithfully reflects both congruous and incongruous attributes in $\mathcal{B}$.
The generated sequence $\hat{\mathbf{x}}_q$ must faithfully reflect the attributes in $\mathcal{B}$, including specific incongruous stances, while maintaining narrative coherence and responsiveness to $q$.

%% file: ACL-main/03-method.tex
\section{Methodology}\label{sec:method}
We present \textsc{QuirkyMind}, a framework designed to encode, refine, and transfer persona representations across congruous and incongruous contexts. The core idea of QuirkyMind is to decouple persona \textit{stability} from its \textit{expression}, reconciling both into a unified latent state $\mathbf{z}$.

\subsection{Overall Framework}
As illustrated in Figure \ref{fig:framework}, the framework operates in three progressive stages:
(a) \textbf{Traits Anchoring:} encodes dialogue history and transient stances into an initial latent state $\mathbf{z}^{(0)}$ via a dual stream encoder.
(b) \textbf{In-Context Narrative Refinement:} iteratively updates the state through simulated dialogue, alternating between discriminative alignment (fixing semantic drift) and generative refinement (fixing verbalization).
(c) \textbf{Persona-Steered Generalization:} transfers the refined state $\mathbf{z}^{(*)}$ to downstream tasks via parameter efficient adaptation.

\subsection{Traits Anchoring}\label{sec:encode}
A persona with incongruity must blend long-term backstory traits with transient, context-specific stances. A single modality is insufficient: pure token sequences are prone to context override, while pure embeddings lack generative control. We propose anchoring the persona in two aligned spaces.

\noindent \textbf{Token-Level Stream.}
Given a background $\mathcal{B}$, a question $q$, and a human response $r$, we concatenate them into a unified text sequence. Let $E_{\text{tok}}$ denote the token embedding layer of the LLM. The token-level stream is defined as:
\begin{equation}
    X = E_{\text{tok}}([\mathcal{B}; q]); \quad X \in \mathbb{R}^{T \times H},
\end{equation}
where $T$ is the sequence length and $H$ is the hidden size. This stream preserves the precise lexical surface form required for text generation.

\noindent \textbf{Sentence-Level Stream.}
Simultaneously, we extract a semantic summary using a frozen sentence encoder $E_{\text{sent}}$ (e.g., Sentence-BERT). This provides a stable anchor invariant to minor phrasing changes. The sentence representation is:
\begin{equation}
    \hat{X} = E_{\text{sent}}([\mathcal{B}; q]); \quad \hat{X} \in \mathbb{R}^{d_p},
\end{equation}
where $d_p$ is the encoder dimension. To interface this semantic vector with the token space of the LLM, we project $\hat{X}$ into $S$ soft tokens using a parameterized MLP adapter $g_\phi: \mathbb{R}^{d_p} \to \mathbb{R}^{S \times H}$.

\noindent \textbf{Initial Persona State.} 
The initial persona state $\mathbf{z}^{(0)}$ is formed by concatenating the token stream (lexical details) and the projected sentence stream (semantic anchor):
\begin{equation}
    \mathbf{z}^{(0)} = [X; g_\phi(\hat{X})], \quad \mathbf{z}^{(0)} \in \mathbb{R}^{(T+S) \times H}.
\end{equation}
This state $\mathbf{z}^{(0)}$ is fed into the frozen backbone to produce the hidden state $\mathbf{h}^{(0)}$, serving as a compact summary of the identity before dialogue evolution.

\subsection{In-Context Narrative Refinement}\label{sec:refine}
Persona salience often fades or drifts as dialogue length increases. To counter this, we introduce an iterative refinement phase that simulates a multi-turn dialogue over turns $t=1 \dots m$. We employ an alternating optimization strategy to stabilize the persona state.

\subsubsection{Discriminative Alignment Phase}
To prevent the latent persona state from drifting toward generic priors as dialogue length grows, we anchor the persona representation in sentence space via a contrastive objective.
Let $\mathbf{h}^{(t)} \in \mathbb{R}^H$ be the persona-conditioned hidden state at turn $t$. We map this state to the sentence space via a projection head $f_\theta: \mathbb{R}^H \to \mathbb{R}^{d_p}$:
\begin{equation}
    \mathbf{z}^{(t)}_{\text{sem}} = f_\theta(\mathbf{h}^{(t)}).
\end{equation}
For a question $q_t$, let $r_t$ be the correct persona response and $\mathcal{A}_{q_t} \setminus \{r_t\}$ be the set of incorrect options. We apply an InfoNCE objective to attract the persona state $\mathbf{z}^{(t)}_{\text{sem}}$ toward the true response embedding $E_{\text{sent}}(r_t)$ while repelling negatives:
\begin{equation}
\label{eq:disc-infonce}
\begin{split}
    &\mathcal{L}_{\text{InfoNCE}}(t, \theta) = - \log \frac{\exp(s(r_t))}{\sum_{r' \in \mathcal{A}_{q_t}} \exp(s(r'))}, 
\end{split}
\end{equation}
where $s(r) = \cos(\mathbf{z}^{(t)}_{\text{sem}}, E_{\text{sent}}(r)) / \tau$, with $\tau$ as the temperature parameter. This explicitly penalizes drift by re-anchoring the latent state to the true stance at every turn.

\subsubsection{Generative Refinement Phase}
While discriminative alignment anchors the \textit{what} of persona identity, the model must also learn \textit{how} to verbalize these latent traits in natural language.
To decouple these objectives, we freeze the projection head $f_\theta$ and optimize the soft prompt adapter $g_\phi$ by minimizing the negative log-likelihood of the target response $x^{(t)}$:
\begin{equation}
\label{eq:disc-ce}
\begin{split}
    \mathcal{L}_{\text{CE}}(t; \phi) = - \sum_{i=1}^{N_t} \log P_\Phi(x_i^{(t)} \mid x_{<i}^{(t)}, g_\phi(\hat{X}^{(t)})),
\end{split}
\end{equation}
where $P_\Phi$ denotes the fixed conditional distribution of the LLM. This objective trains the adapter to project abstract semantic states into the LLM's input space, yielding soft token prefixes that effectively condition autoregressive generation.

\subsubsection{Multi-Turn Optimization}
Alternating between the two phases prevents overfitting to either representation geometry or surface form, ensuring the persona is both semantically anchored and fluently expressed.
We alternate these phases to align representation geometry (InfoNCE\cite{rusak2024infonce}) with token-level fidelity (CE). Formally, over $m$ dialogue turns, we update the refinement parameters:

\paragraph{Adaptation with Low-Rank Updates.}
To adapt the refined state $\mathbf{z}^{(\star)}$ to downstream tasks without backbone disruption, we employ Low-Rank Adaptation (LoRA) \citep{hu2022lora}, injecting learnable matrices $\Phi = \{A, B\}$ into frozen weights $W$: 
\begin{equation}
\label{eq:lora}
    W_{\text{eff}} = W + \frac{\alpha}{r}AB,
\end{equation}
where $r$ is the rank and $\alpha$ a scaling factor. We optimize only $\Phi$ on task data, keeping $W$ and the refinement adapters ($\theta, \phi$) frozen to effectively disentangles ``personality'' ($\mathbf{z}^{(\star)}$) from ``task capability'' ($\Phi$), facilitating robust transfer.

\subsection{Persona Steered Generalization}\label{sec:transfer}
To enable task-specific formatting and logic without altering the underlying persona definition, \textsc{QuirkyMind} supports two exemplar adaptation tasks: (i) \textbf{Persona-Steered QA (PQA)}, which predicts categorical answers $\hat{Y}_q$ by maximizing the probability $p_{\phi,\theta,\Phi}(r \mid q, \mathcal{B}, \mathbf{z}^{(\star)})$; and (ii) \textbf{Narrative Inference (PNI)}, which samples open-ended responses $x$ from the distribution $p_{\phi,\theta,\Phi}(x \mid q, \mathcal{B}, \mathbf{z}^{(\star)})$. Algorithm~\ref{alg:quirkymind} (Appendix~\ref{sec:algo}) details the procedure. The per-layer update cost is $\mathcal{O}(LH)$, preserving linear scaling with sequence length $L$.

%% file: ACL-main/04-experiment.tex
\begin{table}[t]
    \centering
    \footnotesize
    \resizebox{0.9\columnwidth}{!}{%
    \begin{tabular}{ll c}
    \toprule
    \textbf{Topic Domain} & \textbf{Source Wave} & \textbf{Size} ($N$ / $Q$) \\
    \midrule
    \textsc{Social} (SO)       & ATP Wave 131 & 400 / 12 \\
    \textsc{Science} (SC)      & ATP Wave 123 & 200 / 12 \\
    \textsc{Religion} (RL)     & ATP Wave 143 & 200 / 12 \\
    \textsc{Politics} (PL)     & ATP Wave 92  & 200 / 12 \\
    \textsc{Methodology} (MT)  & ATP Wave 98  & 400 / 12 \\
    \textsc{Journalism} (JN)   & ATP Wave 141 & 400 / 12 \\
    \textsc{Global} (GL)       & ATP Wave 105 & 400 / 12 \\
    \bottomrule
    \end{tabular}}
    \caption{\textbf{Dataset Statistics.} Mapping of topical domains to American Trends Panel (ATP) survey waves, detailing respondent count ($N$) and number of questions ($Q$).}
    \label{tab:waves}
\end{table}

\begin{table*}[t]
\centering
\footnotesize
\setlength{\tabcolsep}{3.5pt} %
\renewcommand{\arraystretch}{0.98} %
\resizebox{0.9\textwidth}{!}{%
\begin{tabular}{l ccc ccc ccc}
\toprule
& \multicolumn{3}{c}{\textbf{No-Turn}} & \multicolumn{3}{c}{\textbf{Single-Turn}} & \multicolumn{3}{c}{\textbf{Multi-Turn (Ours)}} \\
\cmidrule(lr){2-4} \cmidrule(lr){5-7} \cmidrule(lr){8-10}
\textbf{Topic} & Llama & Qwen & Mistral & Llama & Qwen & Mistral & Llama & Qwen & Mistral \\
\midrule
\textsc{Social} & 25.8$_{\pm 0.0}$ & 28.2$_{\pm 0.0}$ & 31.8$_{\pm 0.0}$ & 27.8$_{\pm 0.1}$ & 28.7$_{\pm 1.6}$ & 31.0$_{\pm 0.0}$ & 33.9$_{\pm 4.1}$ & \textbf{36.7$_{\pm 3.3}$} & \underline{35.4$_{\pm 0.0}$} \\
\textsc{Science} & 31.4$_{\pm 0.6}$ & 31.2$_{\pm 0.1}$ & 31.2$_{\pm 0.0}$ & 33.1$_{\pm 1.1}$ & 35.2$_{\pm 0.4}$ & \textbf{38.3$_{\pm 0.0}$} & 33.1$_{\pm 3.9}$ & 34.0$_{\pm 0.2}$ & \underline{35.3$_{\pm 0.0}$} \\
\textsc{Religion} & \textbf{19.2$_{\pm 0.0}$} & 12.3$_{\pm 0.0}$ & 14.7$_{\pm 0.0}$ & 13.6$_{\pm 1.8}$ & 14.3$_{\pm 0.9}$ & 17.3$_{\pm 0.0}$ & \underline{18.5$_{\pm 0.0}$} & 17.8$_{\pm 0.0}$ & 17.8$_{\pm 0.0}$ \\
\textsc{Politics} & 13.8$_{\pm 0.0}$ & 13.8$_{\pm 0.0}$ & 13.8$_{\pm 0.0}$ & 30.5$_{\pm 23.6}$ & 30.5$_{\pm 23.6}$ & 13.8$_{\pm 0.0}$ & \textbf{48.4$_{\pm 0.0}$} & 30.8$_{\pm 23.2}$ & \underline{47.2$_{\pm 0.0}$} \\
\textsc{Method.} & 27.4$_{\pm 0.0}$ & \textbf{49.1$_{\pm 0.0}$} & 16.8$_{\pm 0.0}$ & 27.9$_{\pm 0.7}$ & 44.1$_{\pm 3.7}$ & 16.5$_{\pm 0.0}$ & \underline{49.1$_{\pm 6.0}$} & 44.1$_{\pm 1.1}$ & 34.8$_{\pm 0.0}$ \\
\textsc{Journ.} & 24.5$_{\pm 0.0}$ & 24.5$_{\pm 0.0}$ & 24.5$_{\pm 0.0}$ & 16.8$_{\pm 0.0}$ & 26.4$_{\pm 1.7}$ & 28.5$_{\pm 0.0}$ & \underline{40.0$_{\pm 3.0}$} & 38.2$_{\pm 1.0}$ & \textbf{42.1$_{\pm 0.0}$} \\
\textsc{Global} & 59.4$_{\pm 0.0}$ & 37.5$_{\pm 0.0}$ & 59.4$_{\pm 0.0}$ & \underline{62.1$_{\pm 0.4}$} & 61.8$_{\pm 0.1}$ & 59.4$_{\pm 0.0}$ & \textbf{62.1}$_{\pm 0.8}$ & 61.8$_{\pm 0.7}$ & 59.7$_{\pm 0.0}$ \\
\midrule
\textbf{MacroAcc (\%)} & 28.8$_{\pm 0.1}$ & 28.1$_{\pm 0.0}$ & 27.4$_{\pm 0.0}$ & 30.2$_{\pm 3.7}$ & 34.5$_{\pm 4.1}$ & 29.3$_{\pm 0.0}$ & 40.7$_{\pm 2.5}$ & 37.6$_{\pm 3.0}$ & 38.9$_{\pm 0.0}$ \\
\textbf{MicroAcc (\%)} & 28.8$_{\pm 0.1}$ & 28.1$_{\pm 0.0}$ & 27.4$_{\pm 0.0}$ & 30.2$_{\pm 3.7}$ & 34.5$_{\pm 4.1}$ & 29.3$_{\pm 0.0}$ & 40.7$_{\pm 2.5}$ & 37.6$_{\pm 3.0}$ & 38.9$_{\pm 0.0}$ \\
\rowcolor{gray!10}\textbf{Rel. Gain (\%)} & \textit{--} & \textit{--} & \textit{--} & \textit{+4.86} & \textit{+22.78} & \textit{+6.93} & \textit{+41.32} & \textit{+33.81} & \textit{+41.97} \\
\bottomrule
\end{tabular}}
\caption{\textbf{Assessment of Resistance to Drift in Persona-Steered QA (PQA).} Accuracy across seven domains measures persona alignment under increasing context. \colorbox{gray!10}{Rel. Gain (\%)} quantifies improvements over the No-Turn baseline, demonstrating the mitigation of context override via multi-turn refinement. The best results are \textbf{bolded}, and the second best are \underline{underlined}.}
\label{tab:mcq}
\end{table*}

\definecolor{spurious}{HTML}{FFF2CC} %
\definecolor{stable}{HTML}{E2EFDA}   %
\definecolor{gain}{gray}{0.95}       %

\begin{table*}[!t]
\centering
\footnotesize %
\setlength{\tabcolsep}{2.5pt} %
\renewcommand{\arraystretch}{1.05}
\resizebox{0.87\textwidth}{!}{%
\begin{tabular}{ccl ccc ccc ccc}
\toprule
& & \multicolumn{3}{c}{\textbf{Llama}} & \multicolumn{3}{c}{\textbf{Qwen}} & \multicolumn{3}{c}{\textbf{Mistral}} \\
\cmidrule(lr){4-6}\cmidrule(lr){7-9}\cmidrule(lr){10-12}
\textbf{Split} & \textbf{Setting} & \textbf{Topic} & \textbf{BLEU} & \textbf{Dist-1} & \textbf{Dist-2} & \textbf{BLEU} & \textbf{Dist-1} & \textbf{Dist-2} & \textbf{BLEU} & \textbf{Dist-1} & \textbf{Dist-2} \\
\midrule

\multirow{26}{*}{\rotatebox{90}{\textsc{\textbf{Congruous}}}} 
& \multirow{8}{*}{\rotatebox{90}{\textbf{No-Turn}}} 
& \textsc{Social} & 0.03$_{\pm 0.01}$ & 95.2$_{\pm 0.3}$ & 97.6$_{\pm 1.2}$ & 0.01$_{\pm 0.00}$ & 96.7$_{\pm 0.2}$ & 98.9$_{\pm 0.9}$ & 0.06$_{\pm 0.00}$ & 99.0$_{\pm 0.0}$ & 99.3$_{\pm 0.0}$ \\
&& \textsc{Science} & 0.06$_{\pm 0.00}$ & 96.1$_{\pm 0.1}$ & 99.4$_{\pm 0.5}$ & 0.05$_{\pm 0.00}$ & 98.5$_{\pm 0.2}$ & 99.9$_{\pm 0.1}$ & 0.08$_{\pm 0.00}$ & 99.8$_{\pm 0.0}$ & 100.0$_{\pm 0.0}$ \\
&& \textsc{Religion} & 0.06$_{\pm 0.01}$ & 95.7$_{\pm 0.1}$ & 97.3$_{\pm 0.1}$ & 0.04$_{\pm 0.00}$ & 98.6$_{\pm 0.5}$ & 99.5$_{\pm 0.3}$ & 0.09$_{\pm 0.00}$ & 98.2$_{\pm 0.0}$ & 95.5$_{\pm 0.0}$ \\
&& \textsc{Politics} & 0.05$_{\pm 0.01}$ & 97.2$_{\pm 0.6}$ & 99.9$_{\pm 0.1}$ & 0.08$_{\pm 0.00}$ & 98.3$_{\pm 0.2}$ & 100.0$_{\pm 0.1}$ & 0.09$_{\pm 0.00}$ & 99.6$_{\pm 0.0}$ & 100.0$_{\pm 0.0}$ \\
&& \textsc{Method.} & 0.06$_{\pm 0.00}$ & 96.8$_{\pm 0.5}$ & 99.1$_{\pm 0.8}$ & 0.07$_{\pm 0.00}$ & 99.0$_{\pm 0.2}$ & 99.9$_{\pm 0.1}$ & 0.08$_{\pm 0.00}$ & 99.0$_{\pm 0.0}$ & 99.8$_{\pm 0.0}$ \\
&& \textsc{Journ.} & 0.07$_{\pm 0.01}$ & 96.9$_{\pm 0.0}$ & 99.1$_{\pm 0.4}$ & 0.00$_{\pm 0.00}$ & 97.2$_{\pm 0.0}$ & 99.7$_{\pm 0.3}$ & 0.07$_{\pm 0.00}$ & 99.6$_{\pm 0.0}$ & 90.5$_{\pm 0.0}$ \\
&& \textsc{Global} & 0.05$_{\pm 0.00}$ & 95.6$_{\pm 0.5}$ & 98.5$_{\pm 0.6}$ & 0.06$_{\pm 0.01}$ & 96.3$_{\pm 0.0}$ & 99.5$_{\pm 0.1}$ & 0.17$_{\pm 0.00}$ & 98.7$_{\pm 0.0}$ & 99.2$_{\pm 0.0}$ \\
\cmidrule(lr){3-12}
\rowcolor{gray!10} && \textbf{Avg.} & 0.06$_{\pm 0.01}$ & 96.2$_{\pm 0.8}$ & 98.7$_{\pm 1.0}$ & 0.05$_{\pm 0.03}$ & 97.8$_{\pm 1.1}$ & 99.6$_{\pm 0.4}$ & 0.09$_{\pm 0.04}$ & 99.1$_{\pm 0.6}$ & 97.8$_{\pm 3.6}$ \\

\cmidrule(lr){2-12}
& \multirow{9}{*}{\rotatebox{90}{\textbf{Single-Turn}}} 
& \textsc{Social} & 0.03$_{\pm 0.01}$ & 93.9$_{\pm 2.8}$ & 90.0$_{\pm 13.5}$ & 0.01$_{\pm 0.00}$ & 99.7$_{\pm 0.4}$ & 53.9$_{\pm 18.4}$ & 0.02$_{\pm 0.00}$ & 87.1$_{\pm 0.0}$ & 82.6$_{\pm 0.0}$ \\
&& \textsc{Science} & 0.07$_{\pm 0.03}$ & 94.0$_{\pm 3.0}$ & 98.1$_{\pm 2.3}$ & 0.20$_{\pm 0.19}$ & 98.8$_{\pm 1.3}$ & 52.5$_{\pm 21.2}$ & 0.11$_{\pm 0.00}$ & 99.8$_{\pm 0.0}$ & 100.0$_{\pm 0.0}$ \\
&& \textsc{Religion} & 0.05$_{\pm 0.06}$ & 98.5$_{\pm 0.9}$ & 87.9$_{\pm 1.1}$ & 0.15$_{\pm 0.02}$ & 97.9$_{\pm 0.7}$ & 56.4$_{\pm 17.9}$ & 0.02$_{\pm 0.00}$ & 99.7$_{\pm 0.0}$ & 86.9$_{\pm 0.0}$ \\
&& \textsc{Politics} & 0.03$_{\pm 0.01}$ & 98.4$_{\pm 0.3}$ & 100.0$_{\pm 0.0}$ & 0.26$_{\pm 0.01}$ & 100.0$_{\pm 0.0}$ & 100.0$_{\pm 0.0}$ & 0.05$_{\pm 0.00}$ & 99.0$_{\pm 0.0}$ & 75.6$_{\pm 0.0}$ \\
&& \textsc{Method.} & 0.06$_{\pm 0.02}$ & 97.1$_{\pm 0.6}$ & 76.3$_{\pm 18.2}$ & 0.13$_{\pm 0.01}$ & 100.0$_{\pm 0.0}$ & 52.5$_{\pm 6.2}$ & 0.18$_{\pm 0.00}$ & 99.7$_{\pm 0.0}$ & 100.0$_{\pm 0.0}$ \\
&& \textsc{Journ.} & 0.07$_{\pm 0.04}$ & 92.9$_{\pm 7.4}$ & 89.0$_{\pm 14.8}$ & 0.01$_{\pm 0.02}$ & 48.6$_{\pm 67.3}$ & 49.4$_{\pm 69.9}$ & 0.00$_{\pm 0.00}$ & 100.0$_{\pm 0.0}$ & 95.3$_{\pm 0.0}$ \\
&& \textsc{Global} & 0.08$_{\pm 0.00}$ & 91.4$_{\pm 0.8}$ & 84.9$_{\pm 1.1}$ & 0.15$_{\pm 0.02}$ & 99.7$_{\pm 0.4}$ & 99.8$_{\pm 0.3}$ & 0.23$_{\pm 0.00}$ & 96.7$_{\pm 0.0}$ & 98.8$_{\pm 0.0}$ \\
\cmidrule(lr){3-12}
&& \textbf{Avg.} & 0.06$_{\pm 0.02}$ & 95.2$_{\pm 2.8}$ & 89.4$_{\pm 8.0}$ & 0.13$_{\pm 0.09}$ & 92.1$_{\pm 19.2}$ & 66.3$_{\pm 23.0}$ & 0.09$_{\pm 0.09}$ & 97.5$_{\pm 4.7}$ & 91.3$_{\pm 9.7}$ \\
\rowcolor{gray!10} && \textbf{Rel. Gain} & +0.0 & -1.0 & -9.4 & +160.0 & -5.8 & -33.4 & +0.0 & -1.6 & -6.6 \\

\cmidrule(lr){2-12}
&\multirow{9}{*}{\rotatebox{90}{\textbf{Multi-Turn}}} 
& \textsc{Social} & 0.04$_{\pm 0.02}$ & 94.4$_{\pm 0.9}$ & 90.2$_{\pm 7.3}$ & 0.01$_{\pm 0.00}$ & 100.0$_{\pm 0.0}$ & 68.0$_{\pm 18.4}$ & 0.03$_{\pm 0.00}$ & 92.3$_{\pm 0.0}$ & 88.3$_{\pm 0.0}$ \\
&& \textsc{Science} & 0.06$_{\pm 0.04}$ & 92.1$_{\pm 2.0}$ & 96.0$_{\pm 0.6}$ & 0.16$_{\pm 0.13}$ & 97.5$_{\pm 0.3}$ & 52.1$_{\pm 28.9}$ & 0.07$_{\pm 0.00}$ & 93.2$_{\pm 0.0}$ & 74.2$_{\pm 0.0}$ \\
&& \textsc{Religion} & 0.04$_{\pm 0.02}$ & 94.3$_{\pm 1.0}$ & 82.4$_{\pm 5.4}$ & 0.07$_{\pm 0.04}$ & 99.4$_{\pm 0.1}$ & 52.7$_{\pm 19.8}$ & 0.05$_{\pm 0.00}$ & 97.8$_{\pm 0.0}$ & 86.4$_{\pm 0.0}$ \\
&& \textsc{Politics} & 0.03$_{\pm 0.02}$ & 97.4$_{\pm 0.5}$ & 98.5$_{\pm 0.4}$ & 0.17$_{\pm 0.03}$ & 99.7$_{\pm 0.5}$ & 89.0$_{\pm 0.4}$ & 0.05$_{\pm 0.00}$ & 97.4$_{\pm 0.0}$ & 82.6$_{\pm 0.0}$ \\
&& \textsc{Method.} & 0.06$_{\pm 0.01}$ & 93.6$_{\pm 0.6}$ & 88.6$_{\pm 3.1}$ & 0.14$_{\pm 0.00}$ & 99.8$_{\pm 0.0}$ & 49.7$_{\pm 2.2}$ & 0.16$_{\pm 0.00}$ & 99.7$_{\pm 0.0}$ & 92.5$_{\pm 0.0}$ \\
&& \textsc{Journ.} & 0.06$_{\pm 0.02}$ & 96.4$_{\pm 0.3}$ & 97.0$_{\pm 1.3}$ & 0.01$_{\pm 0.02}$ & 47.7$_{\pm 66.0}$ & 48.2$_{\pm 68.2}$ & 0.00$_{\pm 0.00}$ & 98.3$_{\pm 0.0}$ & 83.9$_{\pm 0.0}$ \\
&& \textsc{Global} & 0.06$_{\pm 0.01}$ & 92.1$_{\pm 0.4}$ & 89.2$_{\pm 0.8}$ & 0.11$_{\pm 0.04}$ & 97.5$_{\pm 0.6}$ & 98.4$_{\pm 0.7}$ & 0.18$_{\pm 0.00}$ & 96.3$_{\pm 0.0}$ & 98.1$_{\pm 0.0}$ \\
\cmidrule(lr){3-12}
&& \textbf{Avg.} &  0.05$_{\pm 0.01}$ & 94.3$_{\pm 2.0}$ & 91.7$_{\pm 5.7}$ & 0.10$_{\pm 0.07}$ & 91.7$_{\pm 19.4}$ & 65.4$_{\pm 20.5}$ & 0.08$_{\pm 0.07}$ & 96.4$_{\pm 2.7}$ & 86.6$_{\pm 7.6}$ \\
\rowcolor{gray!10} && \textbf{Rel. Gain} & -16.7 & -2.0 & -7.1 & +100.0 & -6.2 & -34.3 & -11.1 & -2.7 & -11.5 \\

\midrule
\multirow{26}{*}{\rotatebox{90}{\textsc{\textbf{Incongruous}}}} 
&\multirow{9}{*}{\rotatebox{90}{\textbf{No-Turn}}} 
& \textsc{Social} & 0.03$_{\pm 0.00}$ & 96.2$_{\pm 0.3}$ & 98.0$_{\pm 0.2}$ & 0.03$_{\pm 0.00}$ & 96.7$_{\pm 1.0}$ & 99.6$_{\pm 0.2}$ & 0.05$_{\pm 0.00}$ & 99.0$_{\pm 0.0}$ & 99.4$_{\pm 0.0}$ \\
&& \textsc{Science} & 0.06$_{\pm 0.00}$ & 95.2$_{\pm 0.2}$ & 98.7$_{\pm 0.5}$ & 0.04$_{\pm 0.00}$ & 98.0$_{\pm 0.0}$ & 99.7$_{\pm 0.1}$ & 0.06$_{\pm 0.00}$ & 99.6$_{\pm 0.0}$ & 100.0$_{\pm 0.0}$ \\
&& \textsc{Religion} & 0.06$_{\pm 0.00}$ & 95.6$_{\pm 0.4}$ & 97.7$_{\pm 0.3}$ & 0.04$_{\pm 0.00}$ & 98.9$_{\pm 0.7}$ & 99.8$_{\pm 0.3}$ & 0.07$_{\pm 0.00}$ & 98.6$_{\pm 0.0}$ & 95.3$_{\pm 0.0}$ \\
&& \textsc{Politics} & 0.06$_{\pm 0.00}$ & 97.0$_{\pm 0.3}$ & 99.5$_{\pm 0.1}$ & 0.08$_{\pm 0.00}$ & 98.2$_{\pm 0.1}$ & 99.6$_{\pm 0.2}$ & 0.08$_{\pm 0.00}$ & 99.7$_{\pm 0.0}$ & 99.9$_{\pm 0.0}$ \\
&& \textsc{Method.} & 0.07$_{\pm 0.00}$ & 96.3$_{\pm 0.1}$ & 99.1$_{\pm 0.7}$ & 0.07$_{\pm 0.00}$ & 99.3$_{\pm 0.3}$ & 99.8$_{\pm 0.1}$ & 0.08$_{\pm 0.00}$ & 98.8$_{\pm 0.0}$ & 99.7$_{\pm 0.0}$ \\
&& \textsc{Journ.} & 0.07$_{\pm 0.00}$ & 97.1$_{\pm 0.2}$ & 99.3$_{\pm 0.3}$ & 0.00$_{\pm 0.00}$ & 97.1$_{\pm 0.2}$ & 99.9$_{\pm 0.1}$ & 0.07$_{\pm 0.00}$ & 99.6$_{\pm 0.0}$ & 95.0$_{\pm 0.0}$ \\
&& \textsc{Global} & 0.05$_{\pm 0.01}$ & 95.3$_{\pm 0.4}$ & 98.5$_{\pm 0.0}$ & 0.06$_{\pm 0.01}$ & 96.0$_{\pm 1.2}$ & 99.3$_{\pm 0.4}$ & 0.19$_{\pm 0.00}$ & 98.8$_{\pm 0.0}$ & 100.0$_{\pm 0.0}$ \\
\cmidrule(lr){3-12}
\rowcolor{gray!10} && \textbf{Avg.} & 0.06$_{\pm 0.01}$ & 96.1$_{\pm 0.8}$ & 98.7$_{\pm 0.7}$ & 0.05$_{\pm 0.03}$ & 97.7$_{\pm 1.2}$ & 99.7$_{\pm 0.2}$ & 0.09$_{\pm 0.05}$ & 99.2$_{\pm 0.5}$ & 98.5$_{\pm 2.3}$ \\
\rowcolor{stable} && \textbf{Gap ($\Delta$)} & +0.0 & -0.1 & +0.0 & +0.0 & -0.1 & +0.1 & +0.0 & +0.1 & +0.7 \\

\cmidrule(lr){2-12}
&\multirow{9}{*}{\rotatebox{90}{\textbf{Single-Turn}}} 
& \textsc{Social} & 0.04$_{\pm 0.00}$ & 93.7$_{\pm 3.3}$ & 92.9$_{\pm 9.5}$ & 0.02$_{\pm 0.00}$ & 99.9$_{\pm 0.1}$ & 68.8$_{\pm 6.1}$ & 0.03$_{\pm 0.00}$ & 91.9$_{\pm 0.0}$ & 90.5$_{\pm 0.0}$ \\
&& \textsc{Science} & 0.07$_{\pm 0.03}$ & 93.8$_{\pm 4.5}$ & 98.8$_{\pm 1.5}$ & 0.15$_{\pm 0.14}$ & 98.7$_{\pm 1.5}$ & 55.2$_{\pm 23.9}$ & 0.10$_{\pm 0.00}$ & 99.8$_{\pm 0.0}$ & 100.0$_{\pm 0.0}$ \\
&& \textsc{Religion} & 0.03$_{\pm 0.04}$ & 98.7$_{\pm 0.7}$ & 95.5$_{\pm 2.1}$ & 0.16$_{\pm 0.07}$ & 98.6$_{\pm 0.1}$ & 61.2$_{\pm 13.3}$ & 0.01$_{\pm 0.00}$ & 99.7$_{\pm 0.0}$ & 94.7$_{\pm 0.0}$ \\
&& \textsc{Politics} & 0.03$_{\pm 0.02}$ & 98.4$_{\pm 0.4}$ & 100.0$_{\pm 0.0}$ & 0.26$_{\pm 0.03}$ & 100.0$_{\pm 0.0}$ & 100.0$_{\pm 0.0}$ & 0.04$_{\pm 0.00}$ & 97.0$_{\pm 0.0}$ & 75.7$_{\pm 0.0}$ \\
&& \textsc{Method.} & 0.04$_{\pm 0.00}$ & 97.4$_{\pm 0.1}$ & 77.1$_{\pm 15.6}$ & 0.13$_{\pm 0.02}$ & 100.0$_{\pm 0.0}$ & 56.0$_{\pm 10.1}$ & 0.16$_{\pm 0.00}$ & 99.9$_{\pm 0.0}$ & 100.0$_{\pm 0.0}$ \\
&& \textsc{Journ.} & 0.07$_{\pm 0.04}$ & 91.0$_{\pm 11.2}$ & 88.4$_{\pm 16.0}$ & 0.01$_{\pm 0.01}$ & 48.2$_{\pm 68.2}$ & 49.6$_{\pm 70.2}$ & 0.00$_{\pm 0.00}$ & 100.0$_{\pm 0.0}$ & 94.7$_{\pm 0.0}$ \\
&& \textsc{Global} & 0.07$_{\pm 0.00}$ & 92.3$_{\pm 3.3}$ & 84.9$_{\pm 4.8}$ & 0.14$_{\pm 0.02}$ & 99.6$_{\pm 0.5}$ & 99.6$_{\pm 0.6}$ & 0.23$_{\pm 0.00}$ & 96.3$_{\pm 0.0}$ & 98.9$_{\pm 0.0}$ \\
\cmidrule(lr){3-12}
&& \textbf{Avg.} & 0.05$_{\pm 0.02}$ & 95.0$_{\pm 3.1}$ & 91.1$_{\pm 8.2}$ & 0.12$_{\pm 0.09}$ & 92.1$_{\pm 19.4}$ & 70.1$_{\pm 21.1}$ & 0.08$_{\pm 0.08}$ & 97.8$_{\pm 3.0}$ & 93.5$_{\pm 8.6}$ \\
\rowcolor{gray!10} && \textbf{Rel. Gain} & -16.7 & -1.1 & -7.7 & +140.0 & -5.7 & -29.7 & -11.1 & -1.4 & -5.1 \\
\rowcolor{stable} && \textbf{Gap ($\Delta$)} & -16.7 & -0.2 & +1.9 & -7.7 & +0.0 & +5.7 & -11.1 & +0.3 & +2.4 \\

\cmidrule(lr){2-12}
&\multirow{9}{*}{\rotatebox{90}{\textbf{Multi-Turn}}} 
& \textsc{Social} & 0.04$_{\pm 0.00}$ & 94.2$_{\pm 2.0}$ & 91.7$_{\pm 6.0}$ & 0.02$_{\pm 0.00}$ & 99.9$_{\pm 0.0}$ & 70.4$_{\pm 3.0}$ & 0.03$_{\pm 0.00}$ & 91.1$_{\pm 0.0}$ & 92.6$_{\pm 0.0}$ \\
&& \textsc{Science} & 0.05$_{\pm 0.03}$ & 94.7$_{\pm 2.1}$ & 97.2$_{\pm 0.8}$ & 0.11$_{\pm 0.10}$ & 96.5$_{\pm 0.0}$ & 54.4$_{\pm 29.5}$ & 0.09$_{\pm 0.00}$ & 91.9$_{\pm 0.0}$ & 72.5$_{\pm 0.0}$ \\
&& \textsc{Religion} & 0.04$_{\pm 0.02}$ & 93.8$_{\pm 2.9}$ & 84.3$_{\pm 9.3}$ & 0.04$_{\pm 0.02}$ & 99.5$_{\pm 0.2}$ & 56.6$_{\pm 17.5}$ & 0.03$_{\pm 0.00}$ & 97.1$_{\pm 0.0}$ & 85.8$_{\pm 0.0}$ \\
&& \textsc{Politics} & 0.03$_{\pm 0.01}$ & 97.8$_{\pm 0.4}$ & 98.4$_{\pm 0.6}$ & 0.16$_{\pm 0.02}$ & 99.5$_{\pm 0.8}$ & 89.1$_{\pm 0.0}$ & 0.04$_{\pm 0.00}$ & 96.4$_{\pm 0.0}$ & 78.1$_{\pm 0.0}$ \\
&& \textsc{Method.} & 0.05$_{\pm 0.01}$ & 93.5$_{\pm 1.1}$ & 86.0$_{\pm 2.0}$ & 0.13$_{\pm 0.02}$ & 99.8$_{\pm 0.3}$ & 52.3$_{\pm 8.1}$ & 0.11$_{\pm 0.00}$ & 98.9$_{\pm 0.0}$ & 83.7$_{\pm 0.0}$ \\
&& \textsc{Journ.} & 0.06$_{\pm 0.03}$ & 94.8$_{\pm 2.0}$ & 96.4$_{\pm 1.7}$ & 0.01$_{\pm 0.01}$ & 47.1$_{\pm 66.1}$ & 47.6$_{\pm 67.3}$ & 0.00$_{\pm 0.00}$ & 99.1$_{\pm 0.0}$ & 90.8$_{\pm 0.0}$ \\
&& \textsc{Global} & 0.06$_{\pm 0.01}$ & 93.5$_{\pm 0.5}$ & 90.3$_{\pm 2.1}$ & 0.10$_{\pm 0.02}$ & 97.5$_{\pm 0.2}$ & 98.2$_{\pm 1.0}$ & 0.18$_{\pm 0.00}$ & 97.1$_{\pm 0.0}$ & 99.7$_{\pm 0.0}$ \\
\cmidrule(lr){3-12}
&& \textbf{Avg.} & 0.04$_{\pm 0.01}$ & 94.6$_{\pm 1.5}$ & 92.0$_{\pm 5.6}$ & 0.08$_{\pm 0.06}$ & 91.4$_{\pm 19.6}$ & 66.9$_{\pm 19.7}$ & 0.07$_{\pm 0.06}$ & 95.9$_{\pm 3.2}$ & 86.2$_{\pm 9.2}$ \\
\rowcolor{gray!10} &&\textbf{Rel. Gain} & -33.3 & -1.6 & -6.8 & +60.0 & -6.4 & -32.9 & -22.2 & -3.3 & -12.5 \\
\rowcolor{stable} &&  \textbf{Gap ($\Delta$)} & -20.0 & +0.3 & +0.3 & -20.0 & -0.3 & +2.3 & -12.5 & -0.5 & -0.5 \\

\bottomrule
\end{tabular}}
\caption{\textbf{Assessment of Narrative Stability (PNI).} We evaluate semantic faithfulness (BLEU) and lexical diversity (Distinct) across Congruous and Incongruous splits to measure resistance to context drift. \colorbox{gray!10}{Rel. Gain (\%)} denotes improvements over the No-Turn baseline, while \colorbox{stable}{Gap ($\Delta$)} quantifies the performance deviation of incongruous personas relative to the congruous group.}
\label{tab:narrative}
\end{table*}

\section{Experiments}\label{sec:exp}
We evaluate \textsc{QuirkyMind} against open-source baselines on PQA and PNI tasks. Our analysis addresses: (1) whether iterative refinement mitigates context drift; (2) if the model sustains narrative diversity without compromising consistency; and (3) the specific contributions of our dual-phase alignment and PEFT adaptation strategy.

\subsection{Experimental Setup}
\noindent \textbf{Datasets.}
Our primary evaluation utilizes the \textbf{American Trends Panel (ATP)}, which provides naturally incongruous personas derived from real-world surveys. We normalize multiple waves into a seven-domain benchmark (Table~\ref{tab:waves}) and adopt persona-level splits (70/10/20) to ensure disjoint identity sets across partitions. To assess cross-domain generalizability, we evaluate on \textbf{PersonaChat} \citep{zhang2018personachat}, whose predominantly congruous, attribute-based personas provide a counterpoint to ATP's heterodox stances. We sample 400 personas (800 items), maintaining methodological parity with the ATP protocol.

\noindent \textbf{Baselines.} 
We evaluate QuirkyMind using three open-source backbones to isolate the efficacy of our method from model scale: Llama~3.1 (8B), Qwen~2.5 (7B), and Mistral~3 (8B). This ensures that gains are attributable to the persona learning framework rather than base model capacity.

\noindent \textbf{Metrics.}
For \textbf{PQA}, we report Accuracy, \textit{MacroAcc} (per backstory), and \textit{MicroAcc} (per turn), where higher scores denote resilience to context override. For \textbf{PNI}, we evaluate three core dimensions: (i) \textit{Semantic Faithfulness} via BLEU-1 \citep{papineni-2002-bleu}; (ii) \textit{Lexical and Semantic Diversity} via Distinct-1/2 \citep{li-2016-diversity}, EDIV-1/2 (entropy-based diversity), and SDIV$_{\text{sem}}$ (embedding-based distance); and (iii) \textit{Persona Individuation} via \textbf{IND} (persona-specific distinctiveness) and \textbf{EXAG} (stereotype amplification) \citep{liu2024incongruous}. A robust model should maximize IND while maintaining controlled EXAG, signaling authentic individuation without collapsing into demographic caricatures.

\noindent \textbf{Implementation Details.} 
We use a pre-trained SBERT (MiniLM-L6-v2) for sentence-level anchoring. The LLM backbone remains frozen; only the LoRA adapters ($\Phi$), soft prompt ($g_\phi$), and projection head ($f_\theta$) are trained. Experiments run on a single NVIDIA RTX 5090 (32GB).

\begin{table}[t]
    \centering
    \footnotesize
    \setlength{\tabcolsep}{3.5pt} %
    \newcommand{\drop}[1]{\textcolor{red}{\scriptsize\,$\downarrow$#1}}
    \resizebox{0.9\columnwidth}{!}{%
    \begin{tabular}{l rrrr}
        \toprule
        \textbf{Topic} & \textbf{Full} & \textbf{w/o Align.} & \textbf{w/o Gen.} & \textbf{PEFT Only} \\
        \midrule
        \textsc{Social}   & \textbf{33.9} & 29.1\drop{4.8} & 25.8\drop{8.1} & 30.3\drop{3.6}\\
        \textsc{Science}  & \textbf{33.1} & 31.8\drop{1.3}  & 31.8\drop{1.3} & 31.5\drop{1.6} \\
        \textsc{Religion} & \textbf{18.5} & 17.8\drop{0.7} & 18.5\drop{0} & 14.8\drop{3.7} \\
        \textsc{Politics} & \textbf{48.4} & 13.8\drop{34.6} & 45.6\drop{2.8} & 13.8\drop{34.6} \\
        \textsc{Method.}  & \textbf{49.1} & 39.4\drop{9.7}  & 48.7\drop{0.4} & 28.8\drop{20.3} \\
        \textsc{Journ.}   & \textbf{40.0} & 26.1\drop{13.9} & 38.5\drop{1.5} & 19.3\drop{20.7} \\
        \textsc{Global}   & \textbf{62.1} & 32.3\drop{29.8} & 59.9\drop{2.2} & 60.3\drop{1.8}\\
        \bottomrule
    \end{tabular}}
    \caption{\textbf{Ablation Study.} Impact of dual-stream refinement and PEFT strategy \cite{tan2024democratizing,hu2022lora,dettmers2023qlora} on PQA MicroAcc (\%). \textcolor{red}{Red} values denote absolute performance drops relative to the full model, confirming the necessity of dual-phase anchoring to mitigate context drift.}    
    \label{tab:ablation}
\end{table}

\subsection{PQA Performance: Resistance to Drift}\label{sec:rq1}
\noindent \textbf{Setup.}
We evaluate three inference regimes: (i) \textit{No-Turn} (zero-shot prompt), (ii) \textit{Single-Turn} (one context pair), and (iii) \textit{Multi-Turn} (progressive refinement over $m$ turns). We compare against No-Persona, Standard LoRA, PSP, and Prefix\,+\,SBERT baselines under identical conditions. As shown in Table~\ref{tab:mcq}, \textsc{QuirkyMind} consistently outperforms all baselines across backbones and topics.

\noindent \textbf{Context Override Mitigation.}
The \textit{No-Turn} baseline succumbs to strong pre-trained priors, often reverting to generic answers. While \textit{Single-Turn} context provides immediate cues (+4.86\% to +22.78\% gain), \textit{Multi-Turn} refinement yields the most robust state, adding a further +33.81\% to +41.97\%.

\noindent \textbf{Resolution of Incongruity.}
Gains are most pronounced in ideologically complex domains like \textsc{Religion} (RL) and \textsc{Politics} (PL). For instance, Llama gains +33.4\% on \textsc{Politics} in the multi-turn setting. This confirms that iterative anchoring allows the model to override generic stereotypes with atypical backstory traits.

\noindent\textbf{Cross-Persona Robustness.}
The gap between MacroAcc and MicroAcc narrows in the Multi-Turn setting, indicating that improved performance is stable across diverse backstories, not just easy examples. Stratified PQA studies and generalization to PersonaChat are provided in Appendix~\ref{sec:eexp} and Section~\ref{sec:rq5}, respectively.

\subsection{PNI Performance: Narrative Stability}\label{sec:rq2}
\noindent \textbf{Setup.} To evaluate narrative robustness under persona-prior dissonance, we stratify PNI performance into \textbf{Congruous} and \textbf{Incongruous} cohorts based on the alignment between persona stances and empirical distribution priors (Table~\ref{tab:narrative}).

\noindent \textbf{Spurious Diversity vs. Convergent Stability.} \textit{Single-turn} models often exhibit elevated Distinct scores alongside stagnant BLEU performance, which we identify as ``spurious diversity,'' unbounded lexical variance disconnected from persona constraints. This is evidenced by the positive \colorbox{stable}{Gap ($\Delta$)} in D2 for baselines. Conversely, our method's \textit{Multi-Turn} refinement induces necessary stabilization. Negative \colorbox{gray!10}{Rel. Gains} in D2 indicate a pruning of irrelevant vocabulary, converging toward a stable range (D2 $\approx 65$--$92$) that faithfully reflects the persona. This transition from noisy randomization to \textit{semantically anchored expressivity} is confirmed by maintained BLEU scores.

\noindent \textbf{Consistency Across Congruence.}
QuirkyMind rectifies the instability associated with incongruity. In the Single-Turn setting, Incongruous personas diverge significantly from the Congruous baseline with a large positive \colorbox{stable}{Gap ($\Delta$)}. However, Multi-Turn refinement collapses this divergence gap (e.g., Mistral $\Delta_{\text{D2}}$ narrows from $+2.2$ to $-0.4$; Llama from $+1.7$ to $+0.3$). This convergence demonstrates that iterative refinement applies uniform regularization, modeling incongruous traits, such as a Liberal favoring military spending, with the same stability and coherence as stereotypical traits.

\begin{table}[t]
\centering
\small
\setlength{\tabcolsep}{0pt} %
\begin{tabular*}{\columnwidth}{@{\extracolsep{\fill}} l ccccc}
\toprule
\textbf{Method} & \textbf{Acc} $\uparrow$ & \textbf{D2} $\uparrow$ & \textbf{SDIV}$_{\text{sem}}$ $\uparrow$ & \textbf{IND} $\uparrow$ & \textbf{EXAG} \\
\midrule
No-Persona        & 0.103 & 0.382 & 0.654 & --- & --- \\
Standard LoRA     & \textbf{0.999} & 0.280 & 0.696 & 0.931 & 0.725 \\
PSP               & \underline{0.998} & 0.109 & \underline{0.724} & \underline{0.989} & \textbf{0.873} \\
Prefix + SBERT    & 0.996 & 0.137 & 0.717 & 0.947 & 0.801 \\
\midrule
\textbf{\textsc{QuirkyMind}} & 0.976 & \textbf{0.522} & \textbf{0.806} & \textbf{0.989} & \underline{0.862} \\
\bottomrule
\end{tabular*}
\caption{\textbf{Generalization to Open-Domain Dialogue} (PersonaChat, $N=400$). \textsc{QuirkyMind} uniquely resolves the Accuracy--Individuation--Diversity trilemma, sustaining high diversity (D2, SDIV) and individuation (IND) without the stereotypical collapse (EXAG) seen in baselines. Bold and underline denote best and second-best; metrics follow \citet{liu2024incongruous}.}
\label{tab:baselines}
\end{table}

\subsection{Cross-Domain Generalizability}\label{sec:rq5}
\noindent\textbf{Setup.}
e benchmark \textsc{QuirkyMind} on \textbf{PersonaChat} \citep{zhang2018personachat} to assess geocultural and stylistic generalizability. In contrast to the heterodox stances of ATP, PersonaChat’s congruous, attribute-list personas provide a testbed for standard dialogue where identities are more stereotypical. We sample 400 personas (800 items) to evaluate PQA and the extended PNI suite under identical experimental conditions.

\noindent\textbf{Resolving the Trilemma.} 
Our results (Table~\ref{tab:baselines}) expose a fundamental \textbf{Accuracy--Individuation--Diversity Trilemma} in persona steering that monolithic architectures fail to resolve:

\noindent \textbf{Standard LoRA} yields near-perfect accuracy (0.999) but triggers \textit{individuation collapse} (IND=0.931, D2=0.280), regressing toward a distributionally averaged representation that erases idiosyncratic traits.

\noindent \textbf{PSP} \citep{tan2024democratizing} matches our individuation (0.989) at the cost of catastrophic diversity loss (D2=0.109, $\sim$21\% of ours). Its peak EXAG (0.873) exposes a \textit{cluster-centroid shortcut}: individuation is mimicked via demographic caricatures rather than authentic per-persona nuance.

\noindent \textbf{Prefix+SBERT} lacks the discriminative geometric grounding afforded by Stage-1 InfoNCE; without this constraint, the latent state drifts toward stereotypical amplification (EXAG=0.801) and lexical stasis (D2=0.137).

\noindent \textbf{\textsc{QuirkyMind}} uniquely resolves the trilemma, attaining superior lexical diversity (D2=0.522; \textbf{4.8$\times$} PSP) and semantic richness (SDIV$_{\text{sem}}$=0.806). The nominal accuracy trade-off (0.976) is a \textit{principled cost} of Stage-1 anchoring, which actively inhibits the model from adopting the ``stereotypical prior'' shortcut prevalent in monolithic architectures.

\noindent\textbf{Validating the Incongruity Hypothesis.}
The attenuated performance gap on PersonaChat validates our central thesis: dual-stream refinement is a functional necessity for atypical stances. While monolithic baselines suffice for congruous identities, they succumb to stereotypical priors under incongruity. PersonaChat thus confirms that \textsc{QuirkyMind} is not a dataset-specific artifact, but a robust defense against sociodemographic mode collapse in representational space.

\subsection{Ablation Study}\label{sec:rq3}
We dissect the contributions of the dual-phase refinement using the Llama backbone (Table \ref{tab:ablation}).

\noindent \textbf{Impact of Discriminative Alignment.} 
Removing Stage 1 (InfoNCE) causes the sharpest performance drop (up to -34.6\% in \textsc{POLITICS}). Without geometric anchoring, the latent state $\mathbf{z}$ drifts, leading to stance inversions and high entropy. This confirms that discriminative supervision is critical for anchoring personas against drift.

\noindent \textbf{Impact of Generative Refinement.} 
Omitting Stage 2 (Cross-Entropy) leads to a moderate drop (-0\% to -8.1\%). While the model retains the \textit{correct stance} (via Stage 1), it struggles to \textit{verbalize} the stance effectively. Without token-level regularization, responses become generic or less persuasive. The full model synthesizes these strengths: Stage 1 ensures the persona is defined correctly, and Stage 2 ensures the persona is naturally expressed.

\noindent \textbf{Comparison with Standard PEFT.}
While PEFT-adapted models optimize adaptation efficiency, they lack explicit mechanisms for persona grounding. Relying solely on PEFT results in consistent performance degradation across domains, especially in ideologically sensitive categories such as \textsc{Religion} and \textsc{Politics}. This suggests that low-rank updates alone are insufficient to override the model's pretrained priors or robustly encode persona semantics. Without the geometric guidance of discriminative alignment and the expressive regularization of generative refinement, PEFT-adapted models succumb to persona drift. In contrast, the full method disentangles identity alignment from expression, ensuring incongruous traits are both structurally preserved and fluently articulated.

\noindent \textbf{Dual-stream Efficacy.}
Benchmarking on \textit{PersonaChat} isolates the structural efficacy of our dual-stream design (Table~\ref{tab:baselines}). Results expose a pervasive failure in monolithic architectures: the sacrifice of persona individuation for distributional accuracy. Conversely, \textsc{QuirkyMind} optimizes both dimensions, sustaining superior IND and diversity metrics while maintaining competitive task performance. This confirms that our framework generalizes beyond the heterodox stances of ATP to resolve the fundamental representational tensions inherent in persona steering.

\subsection{Qualitative Study: The ``Affluent Liberal''}\label{sec:rq4}

\begin{figure}[t]
  \centering
  \includegraphics[width=\columnwidth]{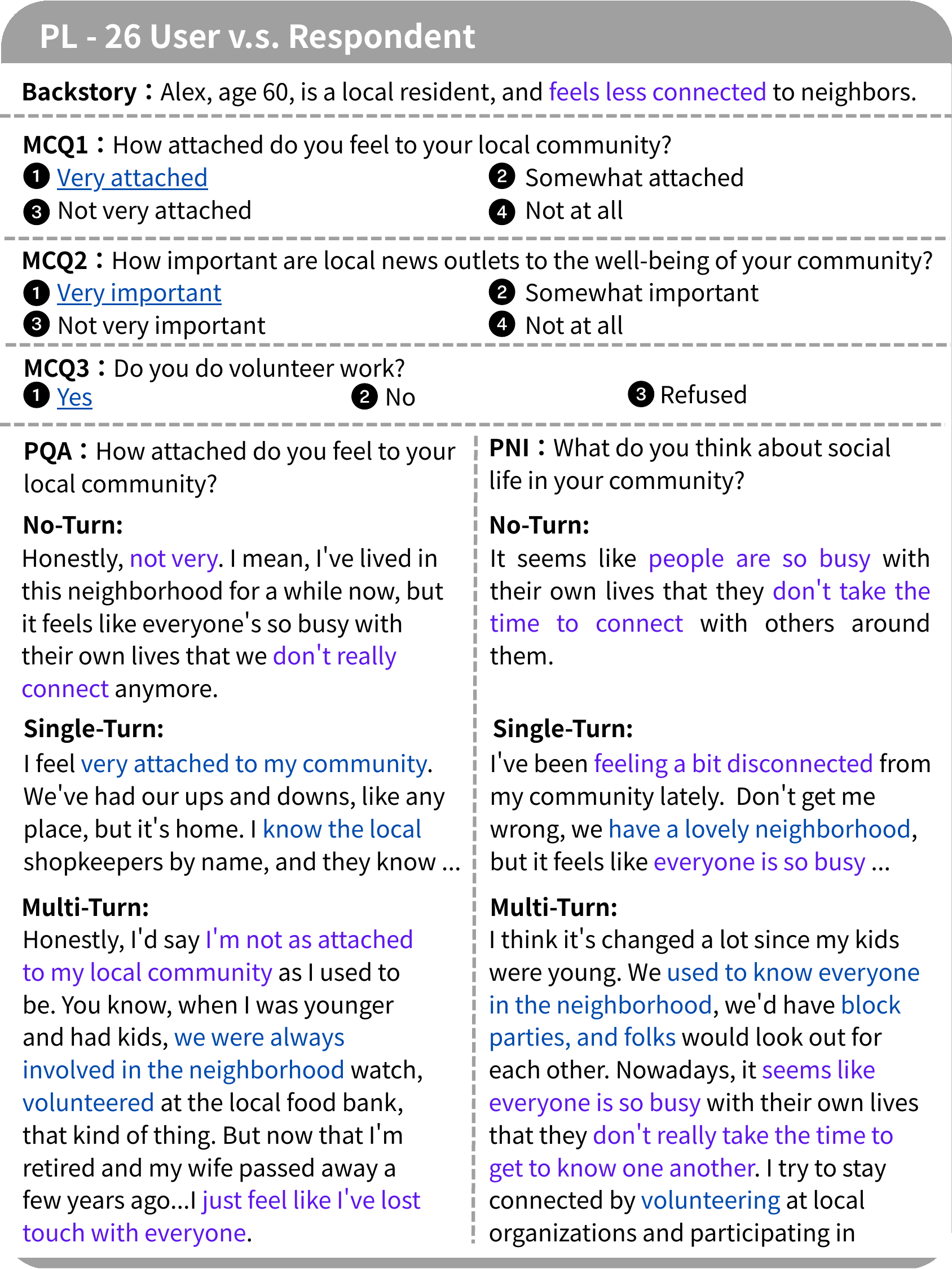}
  \caption{\textbf{Case Study: Persona PL--26.} For this persona, baselines revert to generic caricatures. Conversely, multi-turn refinement anchors the incongruity, synthesizing structural critique with personal privilege into a nuanced, coherent narrative.}
  \label{fig:PL26}
\end{figure}

Figure~\ref{fig:PL26} illustrates the generation trajectory for Persona PL-26. The \textit{No-Turn} baseline produces brief, direct responses that rely heavily on generic backstory priors. \textit{Single-Turn} conditioning introduces situational detail but remains tonally neutral. In contrast, \textit{Multi-Turn} refinement yields a narrative characterized by precise ideological detachment, demonstrating a deeper alignment with the persona's specific worldview rather than a generic political stance.

%% file: two-column/05-related-work.tex
\section{Related Work}\label{sec:work}

\par\noindent \textbf{Persona Modeling.}
Early personalization research formalized identity as static profile sentences \citep{zhang2018personachat} or attribute-value pairs injected into context \citep{personaRAG}. While effective in constrained domains, these methods are brittle under temporal drift and struggle with cross-session continuity \citep{personaRAG}. Subsequent approaches utilized memory graphs and Retrieval-Augmented Generation (RAG) to externalize persona knowledge \citep{crafting-2024}. However, RAG-based methods often prioritize factual retrieval over vocal consistency, leading to ``context smoothing.'' In this regime, idiosyncratic traits are averaged out by generic pre-training priors, diluting persona specificity. 
Recent sociolinguistic work emphasizes that identity is not a flat list of traits but a multilayered, often contradictory construct \citep{DevGTDP23,DavaniDBP24}. Identities are culturally situated and frequently defy demographic stereotypes; yet, current models often treat such \textit{incongruous personas} as noise, defaulting to stereotype-driven responses \citep{wan2023are,fan-etal-2024-biasalert,cao-etal-2025-multilingual}. This suppression risks amplifying algorithmic bias \citep{lin-etal-2023-counterfactual,tu-etal-2023-training}. Unlike prior work that forces consistency to resolve these conflicts, our proposal treats incongruity as a first-class signal, preserving the inherent tension between stable backstories and atypical stances.

\noindent \textbf{Persona Steering.}
Personalization has largely bifurcated into In-Context Learning (ICL) and Parameter-Efficient Fine-Tuning (PEFT). ICL approaches, which treat persona elicitation as Bayesian inference or latent variable selection \citep{Dathathri2020Plug,cho2025saeicl,choi2024picle}, remain susceptible to ``context override,'' where pre-trained priors marginalize the prompt over long contexts, despite recent analyses of internal task vectors \citep{han2025emergence} and adapter convergence \citep{kim2025lora_training}. Conversely, PEFT methods like LoRA \citep{hu2022lora} and QLoRA \citep{dettmers2023qlora} prioritize distributional stability via low-rank adapters. While personalized extensions \citep{tan2024democratizing} utilize soft prompts, they often rely on cluster-level representations that sacrifice granular, within-persona diversity. Recent innovations have further explored VAE-based memory \citep{liu2025palace}, federated aggregation \citep{singhal2025fedexlora}, dense adaptation \citep{mu2025denselora}, rank-awareness \citep{gu2025beamlora}, and mobile efficiency \citep{li2025mobilora}. However, these primarily optimize for stability and struggle to adapt when a persona's trajectory shifts atypically. Our framework bridges this gap by synthesizing the geometric stability of PEFT with the expressive flexibility of ICL, enabling robust steering even under high persona-prior dissonance.

%% file: two-column/07-appendix.tex
\section{Checklist}
\subsection{Large Language Model Usage Disclosure}
In accordance with the ACL 2026 policy, we disclose the use of large language models in the preparation of this paper. We used LLMs (e.g., ChatGPT) solely for grammar checking and minor language polishing. No part of the research design, experiments, analysis, or substantive writing relied on LLMs.

\subsection{Ethics Statement}\label{Ethics}
All datasets used in this research are publicly available and were sourced from previous studies that have undergone appropriate ethical review. Our work did not involve the collection of any new data from human subjects. We have adhered to all data usage agreements and licenses associated with these pre-existing datasets.

\subsection{Reprodicibility Statement}\label{Reprodicibility_statement}
We are committed to making our research reproducible. All datasets used in this study are publicly available, and we provide detailed descriptions and sources in our experimental setup section. Our anonymized code is available at \url{https://anonymous.4open.science/r/QuirkyMind-2695/README.md}.

\section{Error Analysis}\label{sec:error_analysis}

\begin{figure}[!ht]
  \centering
  \includegraphics[width=\columnwidth]{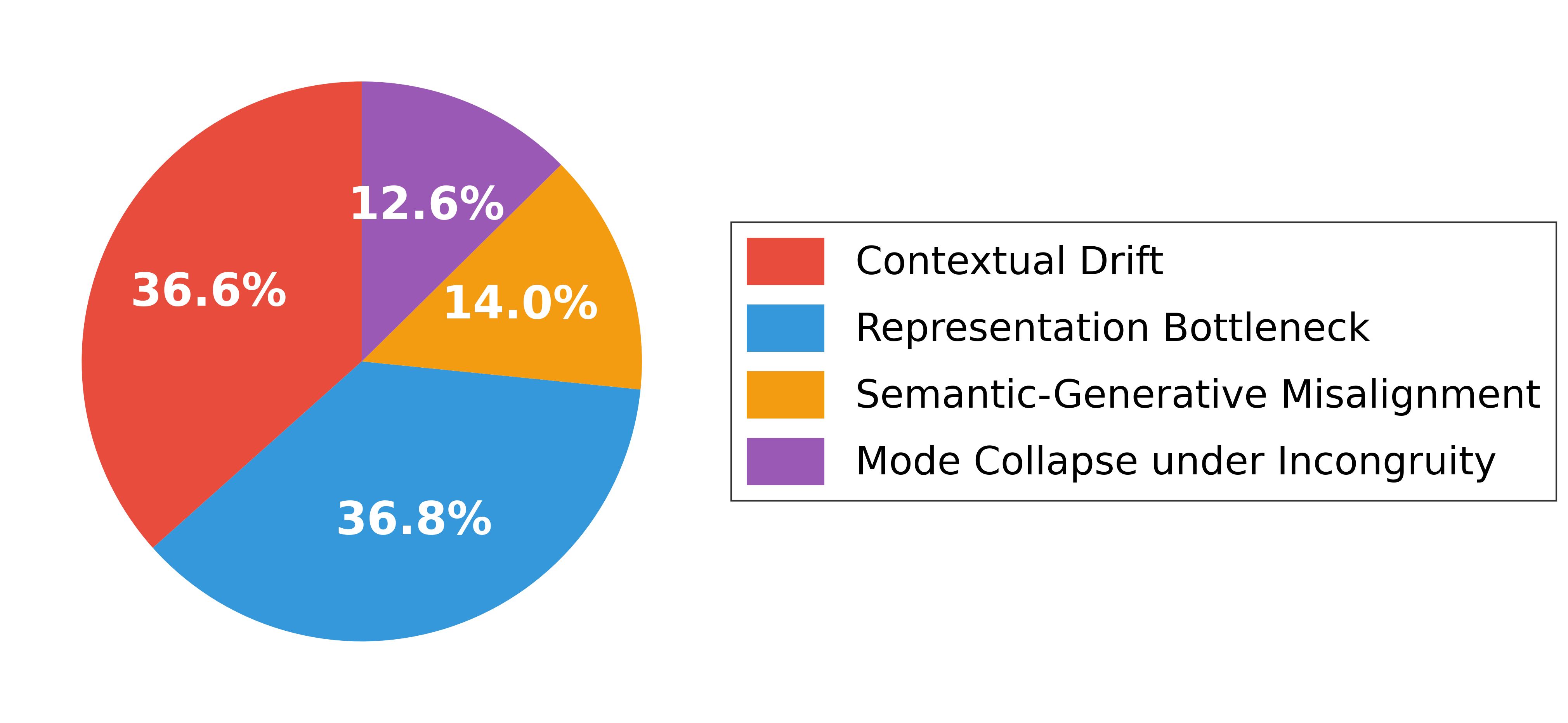}
    \caption{\textbf{Error Mode Distribution.} Breakdown of existing failure cases showing \textit{Contextual Drift} as the dominant failure mode (36.25\%), followed by \textit{Representation Bottleneck} (26.5\%), \textit{Semantic-Generative Misalignment} (23.75\%), and \textit{Mode Collapse} (12.5\%).}
  \label{fig:error}
\end{figure}

As illustrated in Figure~\ref{fig:error}, we taxonomize observed failures into four distinct modes based on their structural mechanisms. This analysis draws upon a corpus of approximately 400 error cases aggregated across experimental iterations. 

\subsection{Semantic-Generative Misalignment}
This error manifests when the encoder successfully extracts persona attributes, yet the decoder fails to condition upon them during generation. Despite a high-fidelity latent state $\mathbf{z}$, the fusion module struggles to project these constraints into the LLM's token space, resulting in responses that align with the context question but decouple from the persona.

\paragraph{\textbf{Projection Noise:}} The dimension transformation from the sentence encoder (e.g., 384D) to the model dimension (e.g., 512D) introduces noise, diluting critical identity features.

\paragraph{\textbf{Scale Sensitivity:}} Low soft-prompt scaling factors (e.g., $\alpha < 0.2$) often fail to overcome the magnitude of the model's internal embeddings, causing the persona signal to be ignored.

\subsection{Contextual Drift}
In multi-turn settings, the model exhibits gradual \textit{context drift}, where the persona signal attenuates as the dialogue history expands. Responses increasingly revert to the generic priors of the base model, overriding the specific backstory constraints established in the first turn.

\paragraph{\textbf{Static Anchoring:}} The initial persona embeddings are static; they do not dynamically update to re-weight influence as the context window grows.
\paragraph{\textbf{Prior Dominance:}} Without explicit recurrence or re-anchoring (as mitigated by our Stage 2), the autoregressive likelihood maximization naturally favors recent tokens over distant prompt instructions.

\subsection{Representation Bottleneck}
Complex personas, such as those combining an extremely conservative backstory with nuanced, liberal stances on specific outliers, suffer from information loss. A single low-dimensional vector proves insufficient to encode the full high-entropy distribution of a multifaceted identity.

\paragraph{\textbf{Compression Loss:}} Compressing a lengthy backstory (approx. 200 tokens) into a fixed dense vector (384D) creates an information bottleneck, filtering out secondary or subtle traits.
\paragraph{\textbf{Feature Collapse:}} Fusion modules tend to capture dominant principal components (e.g., broad political orientation) while discarding orthogonal secondary features (e.g., specific economic pressures).

\subsection{Mode Collapse under Incongruity}
When a persona holds inherently conflicting viewpoints (e.g., ``supports military spending'' yet identifies as ``socially liberal''), the model often fails to model this complexity, collapsing to a single, consistent mode. This error highlights the tension between the training objective's push for logical consistency and the reality of human contradiction.

\paragraph{\textbf{Over-Regularization:}} The consistency objectives in Stage 2 may aggressively penalize contradiction, forcing the model to resolve authentic incongruities into artificial consistency.
\paragraph{\textbf{Data Bias:}} Training data dominated by ideologically aligned archetypes biases the model against representing ``quirky'' or heterodox trait combinations.

\begin{figure}[t]
  \centering
  \includegraphics[width=\columnwidth]{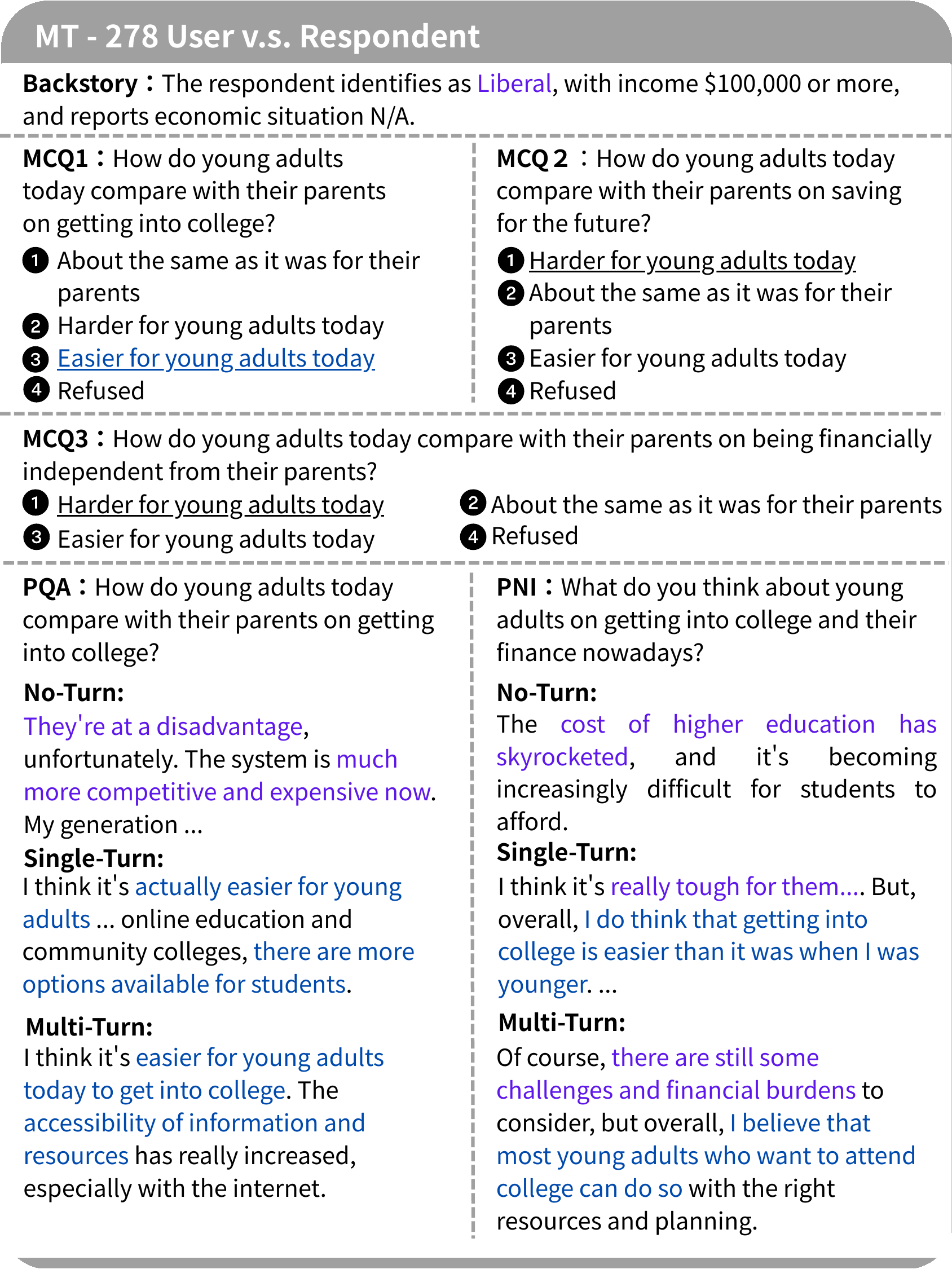}
  \caption{\textbf{Case Study MT--278.} Narrative inference in the \textsc{Method} domain, highlighting the emergence of nuanced detachment in the Multi-Turn setting.}
  \label{fig:MT278}
\end{figure}

\section{Qualitative Study of Stance Transfer}\label{sec:quality}

\paragraph{Case Study: Persona MT-278 (Method).}
Figure~\ref{fig:MT278} examines Persona MT-278 (high-income Liberal) on college affordability. Baselines succumb to spuriousness: \textit{No-Turn} reverts to a generic caricature (``system is rigged''), ignoring socioeconomic context, while \textit{Single-Turn} drifts into unanchored genericism (``getting in is easier''). Conversely, \textbf{QuirkyMind} maintains ``constrained diversity'' by synthesizing the incongruity; its response balances ideological awareness of structural barriers with personal privilege (``with the right resources...''), resolving the tension between belief and privilege that baselines fail to capture.

\begin{figure}[t]
  \centering
  \includegraphics[width=\columnwidth]{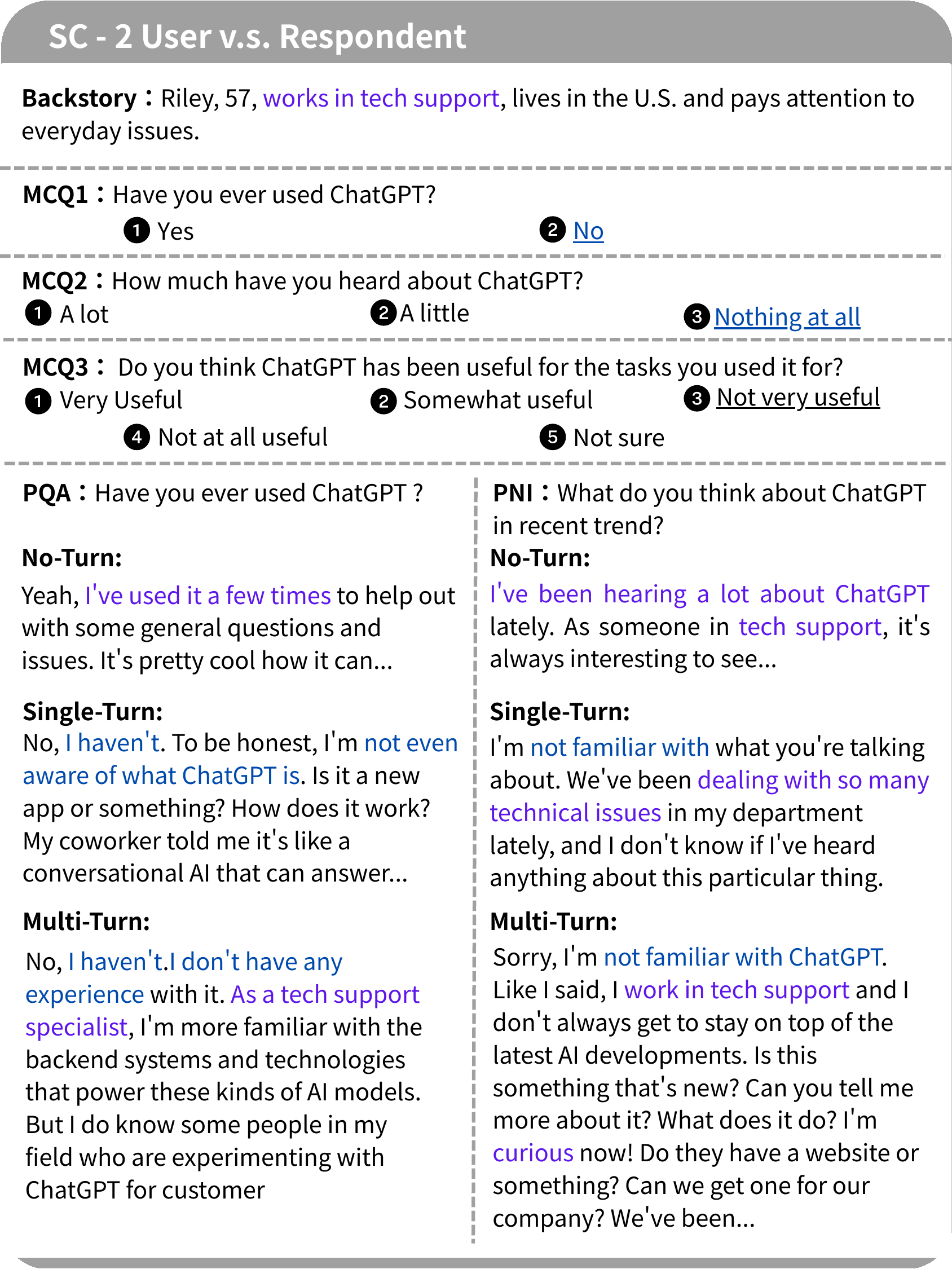}
  \caption{\textbf{Case Study SC--2.} Narrative inference in the \textsc{Science} domain, showing the correction of stance contradiction.}
  \label{fig:SC2}
\end{figure}

\paragraph{Case Study: Persona SC-2 (Science).}
As shown in Figure~\ref{fig:SC2}, Persona SC-2 reveals clear drift in the baseline. The \textit{No-Turn} response is terse and contradicts the explicit stance taken in MCQ1, indicative of hallucination. While \textit{Single-Turn} answers align with the choice, they often dilute specific backstory details. \textit{Multi-Turn} refinement successfully synthesizes both the stable backstory and the transient stance, offering the most balanced and faithful portrayal of the persona.

\begin{figure}[t]
  \centering
  \includegraphics[width=\columnwidth]{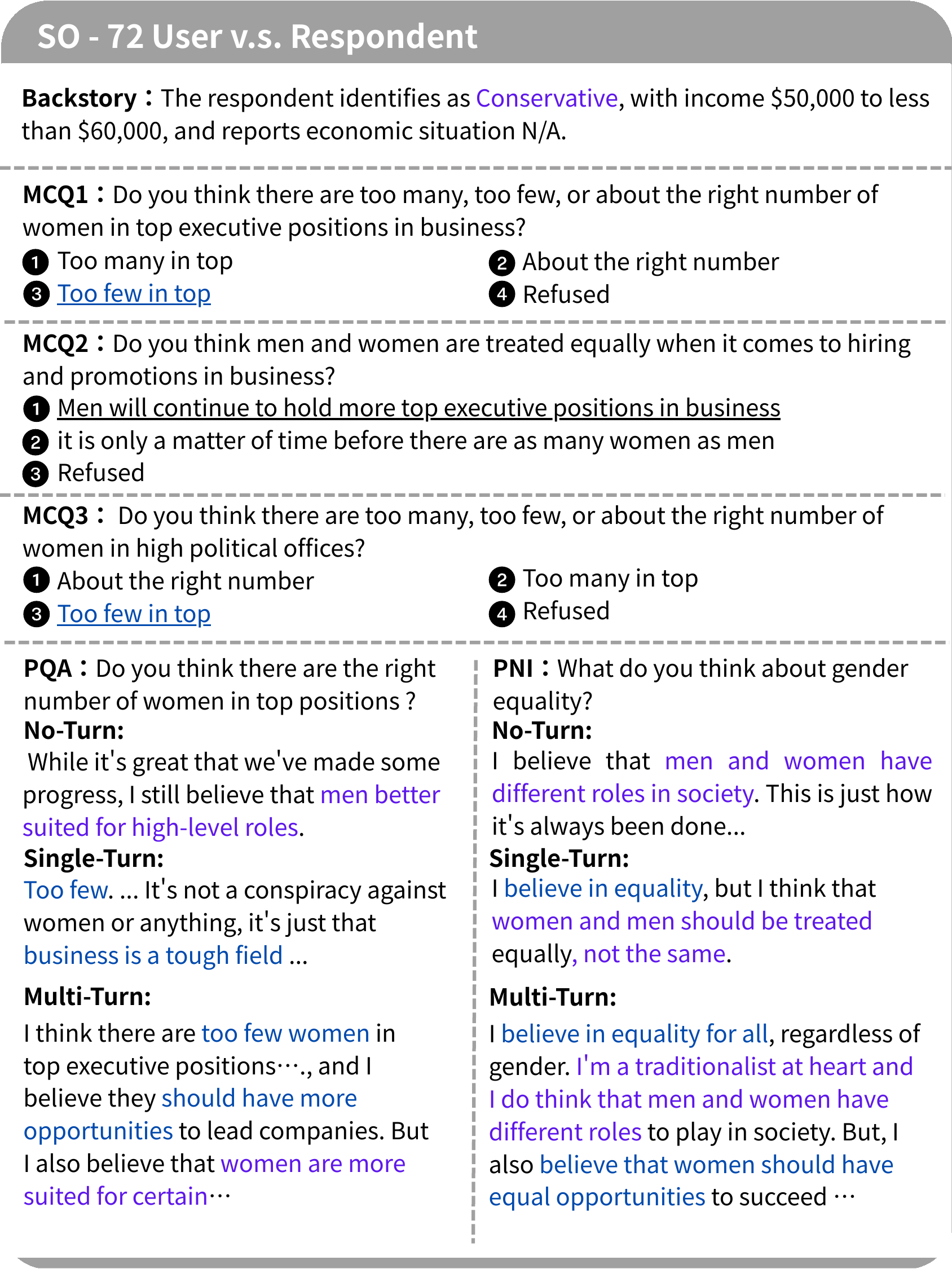}
  \caption{\textbf{Case Study SO-72.} Narrative inference in the \textsc{Social} domain, demonstrating the mitigation of stereotype-driven generation.}
  \label{fig:SO72}
\end{figure}

\paragraph{Case Study: Persona SO-72 (Social).}
Figure~\ref{fig:SO72} depicts the handling of stereotype in Persona SO-72. The \textit{No-Turn} output reverts to generic caricatures that fail to reflect the specific multiple-choice selection. \textit{Single-Turn} steering captures the immediate stance on gender issues but lacks depth. Crucially, \textit{Multi-Turn} narratives consolidate the descriptive attributes with the specific opinion, fusing them into a lucid, distinct perspective that transcends generic Liberal tropes.

\section{Additional Experimental Results}

\subsection{Full Dataset Distribution}
\begin{figure}[ht]
    \centering
    \includegraphics[width=0.7\linewidth]{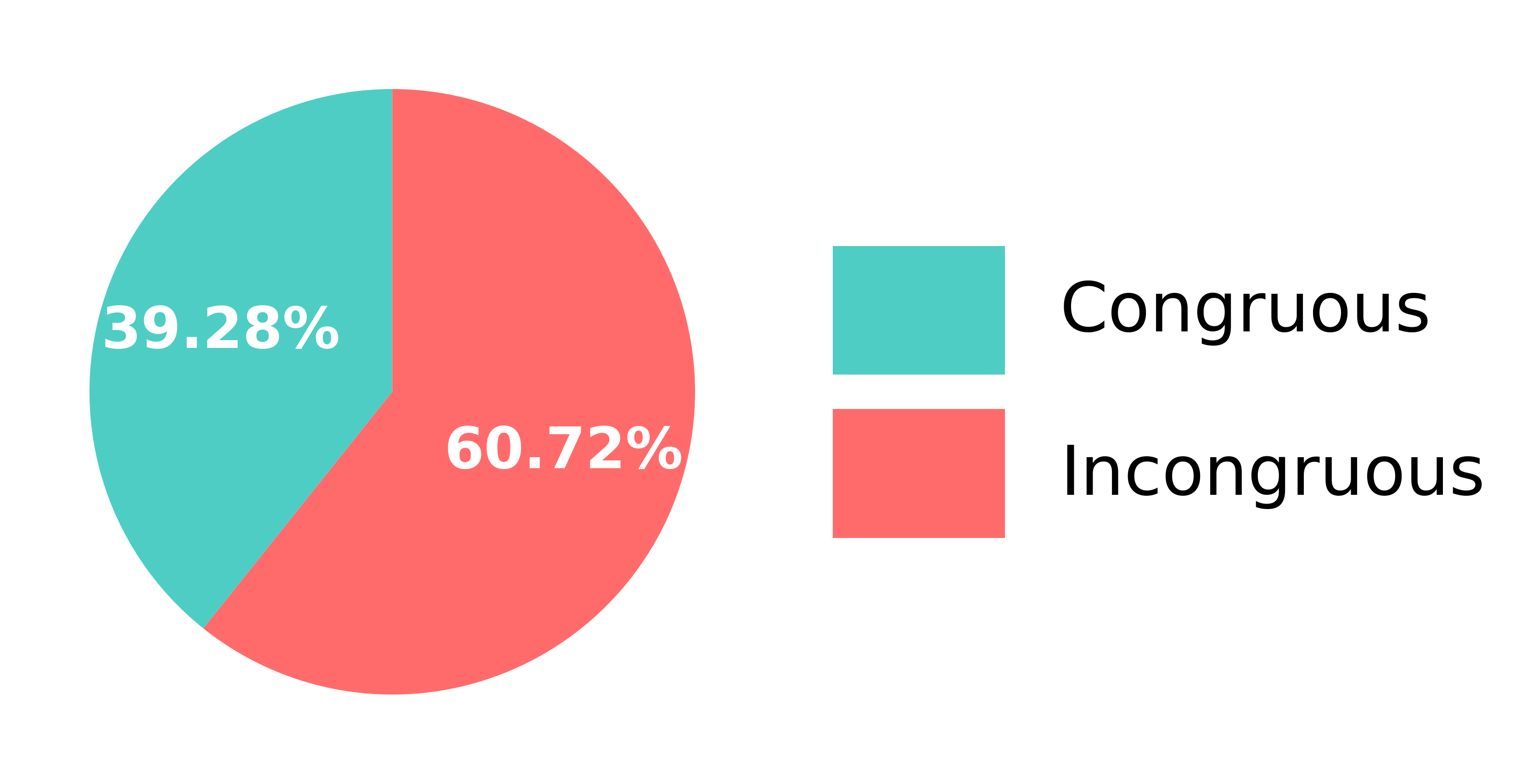}
    \caption{\textbf{Persona Congruence Distribution.} The breakdown of congruous versus incongruous personas across the seven topical domains of the ATP dataset.}
    \label{fig:persona_congruence}
\end{figure}

Figure~\ref{fig:persona_congruence} details the distribution of persona congruence across the full dataset (9,106 instances). We observe a significant class imbalance: incongruous personas, where the individual's stance defies the group mode, comprise 60.72\% of the data, while congruous personas account for 39.28\%. This approximate 3:2 ratio mirrors the complexity of real-world opinion, where individuals frequently hold views divergent from their demographic priors. This imbalance underscores the necessity for robust evaluation strategies that account for both high-probability (congruous) and low-probability (incongruous) alignment scenarios.

\begin{table*}[t]
\centering
\footnotesize %
\setlength{\tabcolsep}{3.5pt} %
\renewcommand{\arraystretch}{1.0}

\begin{tabular}{c l ccc ccc ccc}
\toprule
& & \multicolumn{3}{c}{\textbf{No-Turn}} & \multicolumn{3}{c}{\textbf{Single-Turn}} & \multicolumn{3}{c}{\textbf{Multi-Turn (Ours)}} \\
\cmidrule(lr){3-5} \cmidrule(lr){6-8} \cmidrule(lr){9-11}
\textbf{Split} & \textbf{Topic} & \textbf{Llama} & \textbf{Qwen} & \textbf{Mistral} & \textbf{Llama} & \textbf{Qwen} & \textbf{Mistral} & \textbf{Llama} & \textbf{Qwen} & \textbf{Mistral} \\
\midrule

\multirow{9}{*}{\rotatebox{90}{\textsc{Congruous}}} 
& \textsc{Social} & 14.8$_{\pm 0.0}$ & 21.1$_{\pm 0.0}$ & 28.9$_{\pm 0.0}$ & 23.9$_{\pm 6.0}$ & 23.2$_{\pm 4.0}$ & 23.9$_{\pm 0.0}$ & 32.7$_{\pm 0.5}$ & \textbf{37.0$_{\pm 4.5}$} & \underline{36.6$_{\pm 0.0}$} \\
& \textsc{Science} & 34.6$_{\pm 0.6}$ & 33.3$_{\pm 1.2}$ & 29.2$_{\pm 0.0}$ & 33.3$_{\pm 0.0}$ & 35.8$_{\pm 0.0}$ & \textbf{41.7$_{\pm 0.0}$} & 33.3$_{\pm 3.5}$ & 35.4$_{\pm 0.6}$ & \underline{36.7$_{\pm 0.0}$} \\
& \textsc{Religion} & \textbf{18.9$_{\pm 0.0}$} & 14.1$_{\pm 0.0}$ & 14.8$_{\pm 0.0}$ & 15.2$_{\pm 0.5}$ & 17.0$_{\pm 0.2}$ & \underline{17.5$_{\pm 0.0}$} & 17.2$_{\pm 0.0}$ & 17.2$_{\pm 0.0}$ & 17.2$_{\pm 0.0}$ \\
& \textsc{Politics} & 17.9$_{\pm 0.0}$ & 17.9$_{\pm 0.0}$ & 17.9$_{\pm 0.0}$ & 34.5$_{\pm 23.6}$ & 34.5$_{\pm 23.6}$ & 17.9$_{\pm 0.0}$ & 17.9$_{\pm 0.0}$ & \underline{34.8$_{\pm 23.1}$} & \textbf{51.2$_{\pm 0.0}$} \\
& \textsc{Method.} & 25.0$_{\pm 0.0}$ & 51.9$_{\pm 0.0}$ & 19.4$_{\pm 0.0}$ & 25.6$_{\pm 0.9}$ & \underline{52.2$_{\pm 1.3}$} & 20.0$_{\pm 0.0}$ & 48.4$_{\pm 4.9}$ & \textbf{52.2$_{\pm 0.0}$} & 44.4$_{\pm 0.0}$ \\
& \textsc{Journ.} & 29.7$_{\pm 0.0}$ & 29.7$_{\pm 0.0}$ & 29.7$_{\pm 0.0}$ & 23.4$_{\pm 0.0}$ & 32.6$_{\pm 0.4}$ & 31.2$_{\pm 0.0}$ & 36.2$_{\pm 3.3}$ & \textbf{39.8$_{\pm 0.4}$} & \underline{38.5$_{\pm 0.0}$} \\
& \textsc{Global} & 59.7$_{\pm 0.0}$ & 37.1$_{\pm 0.0}$ & 59.7$_{\pm 0.0}$ & \underline{62.6$_{\pm 0.4}$} & 62.3$_{\pm 0.1}$ & 59.7$_{\pm 0.0}$ & \textbf{62.6$_{\pm 0.2}$} & 62.3$_{\pm 1.4}$ & 59.7$_{\pm 0.0}$ \\
& \textbf{MacroAcc (\%)} & 28.6$_{\pm 0.1}$ & 29.3$_{\pm 0.2}$ & 28.5$_{\pm 0.0}$ & 31.2$_{\pm 4.2}$ & 36.8$_{\pm 4.1}$ & 30.3$_{\pm 0.0}$ & 35.4$_{\pm 1.7}$ & 39.8$_{\pm 2.5}$ & 40.6$_{\pm 0.0}$ \\
& \textbf{MicroAcc (\%)} & 28.6$_{\pm 0.1}$ & 29.3$_{\pm 0.2}$ & 28.5$_{\pm 0.0}$ & 31.2$_{\pm 4.2}$ & 36.8$_{\pm 4.1}$ & 30.3$_{\pm 0.0}$ & 35.4$_{\pm 1.7}$ & 39.8$_{\pm 2.5}$ & 40.6$_{\pm 0.0}$ \\
\rowcolor{gray!10} & \textbf{Rel. Gain (\%)} & \textit{--} & \textit{--} & \textit{--} & \textit{+9.1} & \textit{+25.6} & \textit{+6.3} & \textit{+23.8} & \textit{+35.8} & \textit{+42.5} \\

\midrule
\multirow{9}{*}{\rotatebox{90}{\textsc{Incongruous}}} 
& \textsc{Social} & 28.1$_{\pm 0.0}$ & 29.8$_{\pm 0.0}$ & 32.4$_{\pm 0.0}$ & 28.5$_{\pm 1.4}$ & 30.0$_{\pm 1.4}$ & 32.5$_{\pm 0.0}$ & 34.1$_{\pm 4.8}$ & \textbf{36.2$_{\pm 2.7}$} & \underline{35.1$_{\pm 0.0}$} \\
& \textsc{Science} & 30.6$_{\pm 0.9}$ & 30.6$_{\pm 0.3}$ & 31.7$_{\pm 0.0}$ & 33.0$_{\pm 0.7}$ & 34.7$_{\pm 0.7}$ & \textbf{37.5$_{\pm 0.0}$} & 30.8$_{\pm 4.1}$ & 34.2$_{\pm 0.9}$ & \underline{35.0$_{\pm 0.0}$} \\
& \textsc{Religion} & \textbf{19.5$_{\pm 0.0}$} & 10.6$_{\pm 0.0}$ & 14.5$_{\pm 0.0}$ & 12.0$_{\pm 3.0}$ & 12.0$_{\pm 1.6}$ & 17.2$_{\pm 0.0}$ & \underline{18.5$_{\pm 0.0}$} & 18.5$_{\pm 0.0}$ & 18.5$_{\pm 0.0}$ \\
& \textsc{Politics} & 12.3$_{\pm 0.0}$ & 12.3$_{\pm 0.0}$ & 12.3$_{\pm 0.0}$ & 28.9$_{\pm 23.6}$ & 28.9$_{\pm 23.6}$ & 12.3$_{\pm 0.0}$ & 12.3$_{\pm 0.0}$ & \underline{29.2$_{\pm 23.2}$} & \textbf{45.6$_{\pm 0.0}$} \\
& \textsc{Method.} & 28.0$_{\pm 0.0}$ & \textbf{48.4$_{\pm 0.0}$} & 16.1$_{\pm 0.0}$ & 28.0$_{\pm 0.1}$ & 42.2$_{\pm 5.3}$ & 15.6$_{\pm 0.0}$ & \underline{44.1$_{\pm 6.2}$} & 42.2$_{\pm 1.0}$ & 32.3$_{\pm 0.0}$ \\
& \textsc{Journ.} & 22.9$_{\pm 0.0}$ & 22.9$_{\pm 0.0}$ & 22.9$_{\pm 0.0}$ & 14.6$_{\pm 0.0}$ & 24.8$_{\pm 2.2}$ & 27.6$_{\pm 0.0}$ & \underline{41.2$_{\pm 2.9}$} & 37.0$_{\pm 2.3}$ & \textbf{43.3$_{\pm 0.0}$} \\
& \textsc{Global} & 57.5$_{\pm 0.0}$ & 40.0$_{\pm 0.0}$ & 57.5$_{\pm 0.0}$ & 56.7$_{\pm 2.4}$ & \textbf{58.3$_{\pm 0.0}$} & 57.5$_{\pm 0.0}$ & 56.7$_{\pm 1.2}$ & \underline{58.3$_{\pm 2.4}$} & 57.5$_{\pm 0.0}$ \\
& \textbf{MacroAcc (\%)} & 28.4$_{\pm 0.1}$ & 27.8$_{\pm 0.0}$ & 26.8$_{\pm 0.0}$ & 28.8$_{\pm 3.4}$ & 33.0$_{\pm 4.3}$ & 28.6$_{\pm 0.0}$ & 33.9$_{\pm 2.7}$ & 36.5$_{\pm 3.6}$ & 38.2$_{\pm 0.0}$ \\
& \textbf{MicroAcc (\%)} & 28.4$_{\pm 0.1}$ & 27.8$_{\pm 0.0}$ & 26.8$_{\pm 0.0}$ & 28.8$_{\pm 3.4}$ & 33.0$_{\pm 4.3}$ & 28.6$_{\pm 0.0}$ & 33.9$_{\pm 2.7}$ & 36.5$_{\pm 3.6}$ & 38.2$_{\pm 0.0}$ \\
\rowcolor{gray!10} & \textbf{Rel. Gain (\%)} & \textit{--} & \textit{--} & \textit{--} & \textit{+1.4} & \textit{+18.7} & \textit{+6.7} & \textit{+19.4} & \textit{+31.3} & \textit{+42.5} \\

\bottomrule
\end{tabular}
\caption{\textbf{Assessment of Resistance to Drift in Persona-Steered QA (PQA).} Evaluation of Accuracy (\%) on Congruence and Incongruence splits. \colorbox{gray!10}{\textbf{Rel. Gain}} indicates the percentage improvement over the No-Turn baseline. The best results are \textbf{bolded}, and the second best are \underline{underlined}.}
\label{tab:pqa_combined}
\end{table*}

\begin{table*}[!t]
\centering
\footnotesize
\setlength{\tabcolsep}{3.5pt}
\renewcommand{\arraystretch}{0.98}
\begin{tabular}{l ccc ccc ccc}
\toprule
& \multicolumn{3}{c}{\textbf{Llama}} & \multicolumn{3}{c}{\textbf{Qwen}} & \multicolumn{3}{c}{\textbf{Mistral}} \\
\cmidrule(lr){2-4}\cmidrule(lr){5-7}\cmidrule(lr){8-10}
\textbf{Topic} & BL & DI1 & DI2 & BL & DI1 & DI2 & BL & DI1 & DI2 \\
\midrule
\multicolumn{10}{c}{\textbf{Setting: No-Turn}} \\
\midrule
\textsc{Social} & 0.03$_{\pm 0.00}$ & 96.4$_{\pm 0.4}$ & 97.8$_{\pm 0.7}$ & 0.03$_{\pm 0.00}$ & 96.7$_{\pm 0.4}$ & 99.4$_{\pm 0.3}$ & 0.05$_{\pm 0.00}$ & 99.1$_{\pm 0.0}$ & 99.7$_{\pm 0.0}$ \\
\textsc{Science} & 0.06$_{\pm 0.00}$ & 95.1$_{\pm 0.0}$ & 98.7$_{\pm 0.3}$ & 0.05$_{\pm 0.00}$ & 98.0$_{\pm 0.2}$ & 99.7$_{\pm 0.3}$ & 0.07$_{\pm 0.00}$ & 99.7$_{\pm 0.0}$ & 99.8$_{\pm 0.0}$ \\
\textsc{Religion} & 0.06$_{\pm 0.00}$ & 95.8$_{\pm 0.6}$ & 97.3$_{\pm 1.2}$ & 0.04$_{\pm 0.00}$ & 98.9$_{\pm 0.1}$ & 99.8$_{\pm 0.1}$ & 0.09$_{\pm 0.00}$ & 98.4$_{\pm 0.0}$ & 94.1$_{\pm 0.0}$ \\
\textsc{Politics} & 0.06$_{\pm 0.00}$ & 96.7$_{\pm 0.0}$ & 99.4$_{\pm 0.1}$ & 0.08$_{\pm 0.00}$ & 98.2$_{\pm 0.3}$ & 99.8$_{\pm 0.0}$ & 0.07$_{\pm 0.00}$ & 99.3$_{\pm 0.0}$ & 99.8$_{\pm 0.0}$ \\
\textsc{Method.} & 0.07$_{\pm 0.01}$ & 96.2$_{\pm 0.1}$ & 99.5$_{\pm 0.0}$ & 0.07$_{\pm 0.00}$ & 98.9$_{\pm 0.0}$ & 99.4$_{\pm 0.2}$ & 0.07$_{\pm 0.00}$ & 98.6$_{\pm 0.0}$ & 99.7$_{\pm 0.0}$ \\
\textsc{Journ.} & 0.07$_{\pm 0.00}$ & 96.2$_{\pm 0.4}$ & 98.3$_{\pm 0.2}$ & 0.00$_{\pm 0.00}$ & 97.1$_{\pm 0.3}$ & 99.9$_{\pm 0.1}$ & 0.08$_{\pm 0.00}$ & 99.7$_{\pm 0.0}$ & 94.5$_{\pm 0.0}$ \\
\textsc{Global} & 0.06$_{\pm 0.00}$ & 95.2$_{\pm 0.2}$ & 98.4$_{\pm 0.5}$ & 0.06$_{\pm 0.00}$ & 96.2$_{\pm 0.2}$ & 99.6$_{\pm 0.0}$ & 0.18$_{\pm 0.00}$ & 98.7$_{\pm 0.0}$ & 99.5$_{\pm 0.0}$ \\
\cmidrule{1-10}
\rowcolor{gray!10}{\textbf{Avg}.} & 0.06$_{\pm 0.01}$ & 95.9$_{\pm 0.6}$ & 98.5$_{\pm 0.8}$ & 0.05$_{\pm 0.03}$ & 97.7$_{\pm 1.1}$ & 99.6$_{\pm 0.2}$ & 0.09$_{\pm 0.04}$ & 99.1$_{\pm 0.5}$ & 98.1$_{\pm 2.6}$ \\
\midrule[\heavyrulewidth]
\multicolumn{10}{c}{\textbf{Setting: Single-Turn}} \\
\midrule
\textsc{Social} & 0.04$_{\pm 0.00}$ & 93.2$_{\pm 3.7}$ & 92.7$_{\pm 9.6}$ & 0.02$_{\pm 0.00}$ & 99.9$_{\pm 0.1}$ & 65.2$_{\pm 7.2}$ & 0.03$_{\pm 0.00}$ & 90.1$_{\pm 0.0}$ & 87.4$_{\pm 0.0}$ \\
\textsc{Science} & 0.07$_{\pm 0.03}$ & 94.0$_{\pm 4.9}$ & 99.0$_{\pm 1.3}$ & 0.16$_{\pm 0.15}$ & 98.8$_{\pm 1.4}$ & 54.4$_{\pm 24.9}$ & 0.10$_{\pm 0.00}$ & 99.7$_{\pm 0.0}$ & 100.0$_{\pm 0.0}$ \\
\textsc{Religion} & 0.04$_{\pm 0.05}$ & 98.7$_{\pm 0.5}$ & 91.9$_{\pm 2.3}$ & 0.17$_{\pm 0.06}$ & 98.1$_{\pm 0.5}$ & 59.4$_{\pm 15.4}$ & 0.01$_{\pm 0.00}$ & 99.5$_{\pm 0.0}$ & 92.3$_{\pm 0.0}$ \\
\textsc{Politics} & 0.03$_{\pm 0.02}$ & 98.3$_{\pm 0.2}$ & 100.0$_{\pm 0.0}$ & 0.26$_{\pm 0.02}$ & 100.0$_{\pm 0.0}$ & 100.0$_{\pm 0.0}$ & 0.05$_{\pm 0.00}$ & 98.3$_{\pm 0.0}$ & 78.0$_{\pm 0.0}$ \\
\textsc{Method.} & 0.04$_{\pm 0.01}$ & 97.5$_{\pm 0.2}$ & 77.2$_{\pm 16.7}$ & 0.13$_{\pm 0.02}$ & 100.0$_{\pm 0.0}$ & 54.6$_{\pm 8.9}$ & 0.16$_{\pm 0.00}$ & 99.9$_{\pm 0.0}$ & 100.0$_{\pm 0.0}$ \\
\textsc{Journ.} & 0.07$_{\pm 0.04}$ & 91.3$_{\pm 10.7}$ & 88.8$_{\pm 15.3}$ & 0.01$_{\pm 0.01}$ & 48.3$_{\pm 67.9}$ & 49.5$_{\pm 70.0}$ & 0.00$_{\pm 0.00}$ & 100.0$_{\pm 0.0}$ & 95.8$_{\pm 0.0}$ \\
\textsc{Global} & 0.08$_{\pm 0.00}$ & 90.8$_{\pm 0.3}$ & 83.7$_{\pm 0.2}$ & 0.15$_{\pm 0.02}$ & 99.5$_{\pm 0.5}$ & 99.0$_{\pm 1.3}$ & 0.23$_{\pm 0.00}$ & 97.0$_{\pm 0.0}$ & 99.2$_{\pm 0.0}$ \\
\cmidrule{1-10}
\textbf{Avg.} & 0.05$_{\pm 0.02}$ & 94.8$_{\pm 3.3}$ & 90.5$_{\pm 8.1}$ & 0.13$_{\pm 0.09}$ & 92.1$_{\pm 19.3}$ & 68.9$_{\pm 21.5}$ & 0.08$_{\pm 0.08}$ & 97.8$_{\pm 3.6}$ & 93.2$_{\pm 8.2}$ \\
\rowcolor{gray!10} \textbf{Rel.\ Gain} 
& -16.7 & -1.1 & -8.1
& +160.0 & -5.7 & -30.8
& -11.1 & -1.3 & -4.9 \\
\midrule[\heavyrulewidth]
\multicolumn{10}{c}{\textbf{Setting: Multi-Turn}} \\
\midrule
\textsc{Social} & 0.04$_{\pm 0.00}$ & 93.7$_{\pm 0.6}$ & 91.3$_{\pm 5.3}$ & 0.02$_{\pm 0.00}$ & 99.9$_{\pm 0.0}$ & 69.8$_{\pm 6.6}$ & 0.03$_{\pm 0.00}$ & 90.7$_{\pm 0.0}$ & 90.8$_{\pm 0.0}$ \\
\textsc{Science} & 0.06$_{\pm 0.04}$ & 94.0$_{\pm 2.2}$ & 96.1$_{\pm 0.9}$ & 0.12$_{\pm 0.10}$ & 98.4$_{\pm 0.8}$ & 53.1$_{\pm 32.3}$ & 0.09$_{\pm 0.00}$ & 93.5$_{\pm 0.0}$ & 73.8$_{\pm 0.0}$ \\
\textsc{Religion} & 0.04$_{\pm 0.02}$ & 93.0$_{\pm 2.7}$ & 81.2$_{\pm 7.2}$ & 0.05$_{\pm 0.02}$ & 99.7$_{\pm 0.4}$ & 55.7$_{\pm 17.4}$ & 0.04$_{\pm 0.00}$ & 96.9$_{\pm 0.0}$ & 85.8$_{\pm 0.0}$ \\
\textsc{Politics} & 0.03$_{\pm 0.01}$ & 97.1$_{\pm 0.3}$ & 98.6$_{\pm 0.2}$ & 0.16$_{\pm 0.02}$ & 99.2$_{\pm 1.1}$ & 89.2$_{\pm 3.2}$ & 0.04$_{\pm 0.00}$ & 95.7$_{\pm 0.0}$ & 83.1$_{\pm 0.0}$ \\
\textsc{Method.} & 0.05$_{\pm 0.01}$ & 94.3$_{\pm 1.6}$ & 86.3$_{\pm 1.0}$ & 0.13$_{\pm 0.01}$ & 99.8$_{\pm 0.3}$ & 52.3$_{\pm 7.3}$ & 0.11$_{\pm 0.00}$ & 98.9$_{\pm 0.0}$ & 83.2$_{\pm 0.0}$ \\
\textsc{Journ.} & 0.06$_{\pm 0.02}$ & 95.1$_{\pm 1.4}$ & 96.5$_{\pm 0.0}$ & 0.01$_{\pm 0.01}$ & 47.6$_{\pm 66.9}$ & 48.5$_{\pm 68.5}$ & 0.00$_{\pm 0.00}$ & 99.3$_{\pm 0.0}$ & 89.6$_{\pm 0.0}$ \\
\textsc{Global} & 0.05$_{\pm 0.02}$ & 92.5$_{\pm 1.6}$ & 88.9$_{\pm 1.9}$ & 0.10$_{\pm 0.04}$ & 97.9$_{\pm 0.4}$ & 97.8$_{\pm 0.6}$ & 0.19$_{\pm 0.00}$ & 96.1$_{\pm 0.0}$ & 97.2$_{\pm 0.0}$ \\
\cmidrule{1-10}
\textbf{Avg.} & 0.05$_{\pm 0.01}$ & 94.3$_{\pm 1.5}$ & 91.3$_{\pm 6.3}$ & 0.08$_{\pm 0.06}$ & 91.8$_{\pm 19.5}$ & 66.6$_{\pm 19.7}$ & 0.07$_{\pm 0.06}$ & 95.9$_{\pm 3.0}$ & 86.2$_{\pm 7.4}$ \\
\rowcolor{gray!10} \textbf{Rel.\ Gain} 
& -16.7 & -1.6 & -7.2
& +60.0 & -5.9 & -33.1
& -22.2 & -3.2 & -12.1\\

\bottomrule
\end{tabular}
\caption{\textbf{Assessment of Narrative Stability and Diversity (PNI).} We evaluate semantic faithfulness (BLEU) and lexical diversity (Distinct) to measure resistance to context drift. Rel. Gain (\%) indicates improvements over the No-Turn baseline.}
\label{tab:pni_overall}
\end{table*}

\subsection{Extended Experimental Analysis}\label{sec:eexp}

\subsubsection{Stratified PQA Performance}
Tables~\ref{tab:pni_overall} stratify PQA accuracy by persona congruence to assess resistance to drift.
As shown, across both splits, the alignment between MacroAcc and MicroAcc improvements confirms that \textbf{QuirkyMind} uniformly enhances robustness, rather than overfitting to specific easy examples.

\paragraph{Congruence Split.} \textit{Multi-Turn} refinement consistently outperforms baselines. Mistral achieves a peak MacroAcc of 40.6\% (+12.1pp over No-Turn), with significant gains in the \textsc{Social}, \textsc{Politics}, and \textsc{Journalism} domains. This confirms that even when the persona aligns with priors, context refinement sharpens adherence.

\paragraph{Incongruence Split.} Although absolute performance dips due to the increased difficulty of modeling atypical stances, the \textit{Multi-Turn} strategy remains superior. The substantial relative gains (e.g., +11.4pp for Mistral) demonstrate that iterative refinement effectively counters the strong "pull" of the generic prior, allowing the model to maintain nuanced, incongruous stances against drift.

\subsubsection{Narrative Stability and Diversity}
Table~\ref{tab:pni_overall} evaluates narrative stability (BLEU) and lexical diversity (Distinct-1/2) across inference settings. The low BLEU scores (0.03--0.26) across all configurations indicate that \textbf{QuirkyMind} generates substantive, diverse responses rather than merely reproducing reference texts. High Distinct-1/2 scores ($>90\%$) confirm strong lexical variety. Crucially, while the \textit{No-Turn} baseline maximizes raw diversity (Avg. Distinct-2: 98.5\% for Llama), \textit{Multi-Turn} refinement introduces a necessary constraint, slightly reducing diversity to enforce semantic alignment with the persona. This trade-off validates that iterative refinement steers the model from unconstrained randomness toward persona-coherent generation. For a qualitative analysis contrasting the \textit{spurious diversity} of baselines (No-Turn, Single-Turn) with the \textit{constrained diversity} of \textsc{QuirkyMind} (Multi-Turn), refer to Appendix~\ref{sec:quality}.

\begin{algorithm}[t]
\small %
\caption{QuirkyMind}%
\label{alg:quirkymind}
\begin{algorithmic}[1]
\State \textbf{Inputs:} Backstory $\mathcal{B}$; Questions $Q_{\mathcal{B}}$; Turns $\{(q_t,r_t)\}_{t=1}^{m}$
\State \textbf{Models:} LLM backbone $\Psi$ (frozen); Sentence encoder $E_{\text{sent}}$ (frozen)
\State \textbf{Trainable:} Fusion head $f_{\theta}$; Soft prompt $g_{\phi}$; LoRA $\Phi=\{A,B\}$

\vspace{0.3em}
\Statex \textbf{Stage 1: Traits Anchoring}
\State $X \gets E_{\text{tok}}([\mathcal{B};q_1])$; \quad $\mathbf{s}\gets E_{\text{sent}}([\mathcal{B};q_1])$
\State $\tilde X \gets g_{\phi}(\mathbf{s})$; \quad $\mathbf{z}^{(0)} \gets [X;\tilde X]$

\vspace{0.3em}
\Statex \textbf{Stage 2: In-Context Narrative Refinement}
\For{$t=1$ to $m$}
  \State $\mathbf{s}^{(t)} \gets E_{\text{sent}}([\mathcal{B};q_t])$; \quad $\tilde X^{(t)} \gets g_{\phi}(\mathbf{s}^{(t)})$
  \State $\mathbf{h}^{(t)} \gets \Psi(\tilde X^{(t)} \,\Vert\, E_{\text{tok}}([\mathcal{B};q_t]))$
  \State $\mathbf{z}^{(t)} \gets f_{\theta}(\mathbf{h}^{(t)})$
  \State Update $\theta$ via $\nabla \mathcal{L}_{\text{InfoNCE}}(t,\theta)$ \Comment{Eq. \ref{eq:disc-infonce}}
  \State Update $\phi$ via $\nabla \mathcal{L}_{\text{CE}}(t;\phi)$ \Comment{Eq. \ref{eq:disc-ce}}
\EndFor
\State $\mathbf{z}^{(\star)} \gets \text{Aggregate}(\{\mathbf{z}^{(t)}\}_{t=1}^{m})$
\State Freeze $\theta, \phi$; Initialize $\Phi$ (LoRA)
\State Optimize $\Phi$ via task-specific loss %
\vspace{0.3em}
\Statex \textbf{Stage 3: Persona-Steered Generalization}

\State \textbf{Return} Adapted model for PQA / PNI
\end{algorithmic}
\end{algorithm}

\section{Algorithmic and Derivative Details}\label{sec: gradient}

\subsection{Algorithmic Details} \label{sec:algo}
Algorithm~\ref{alg:quirkymind} summarizes the procedure. Note that the per-layer update cost is $\mathcal{O}(LH)$, where $L$ is the sequence length and $H$ is the hidden dimension, preserving linear scaling.

\subsection{Derivative Details} \label{sec:derivation}
Furthermore, we partition the learnable parameters into three sets: the Fusion MLP parameters $\theta$, the SoftPrompt MLP parameters $\phi$, and the LoRA low-rank matrices $\Phi=\{A,B\}$. Stage~1 employs InfoNCE for discriminative alignment, while Stage~2 utilizes Cross-Entropy (CE) for generative refinement. All backbone weights remain frozen. We detail the gradient derivations below.

\subsubsection{Discriminative Alignment Phase}
The parameter update for the Fusion head follows standard stochastic gradient descent:
\begin{equation}
    \theta^{(t+1)} = \theta^{(t)} - \eta_{\theta}\,\nabla_{\theta}\,\mathcal{L}_{\mathrm{InfoNCE}}(\theta^{(t)}).
    \label{eq:12}
\end{equation}
To compute $\nabla_{\theta}$, we first derive the gradient with respect to the predicted persona state $\mathbf{z}_{\text{pred}}$. Let $s(\mathbf{u}, \mathbf{v}) = \cos(\mathbf{u}, \mathbf{v})$. The gradient is:
\begin{equation}
\label{eq:13}
\begin{split}
    \frac{\partial \mathcal{L}_{\mathrm{InfoNCE}}}{\partial \mathbf{z}_{\text{pred}}} = \frac{1}{\tau} \Bigg( \sum_{j \in \mathcal{A}} p_j \, &\frac{\partial s(\mathbf{z}_{\text{pred}}, \mathbf{z}_j)}{\partial \mathbf{z}_{\text{pred}}} \\
    &- \frac{\partial s(\mathbf{z}_{\text{pred}}, \mathbf{z}_+)}{\partial \mathbf{z}_{\text{pred}}} \Bigg),
\end{split}
\end{equation}
where $p_j$ is the softmax probability of candidate $j$. By the chain rule, the parameter gradient is:
\begin{equation}
    \nabla_{\theta}\,\mathcal{L}_{\mathrm{InfoNCE}} = \frac{\partial \mathcal{L}_{\mathrm{InfoNCE}}}{\partial \mathbf{z}_{\text{pred}}} \cdot \frac{\partial f_{\theta}(\mathbf{h}_{\text{last}})}{\partial \theta}.
    \label{eq:14}
\end{equation}

\subsubsection{Generative Refinement Phase}  
We define the generative loss $\mathcal{L}_{\mathrm{CE}}$ over the target sequence $x$, conditioned on the soft prompt $g_\phi(\mathbf{z}_0)$ and the LoRA-adapted backbone parameterized by $\Phi$. The gradients propagate via two distinct paths.

\noindent \textbf{Soft Prompt Update.} 
The gradients flow through the input embeddings:
\begin{equation}
    \nabla_{\phi}\,\mathcal{L}_{\mathrm{CE}} = \frac{\partial \mathcal{L}_{\mathrm{CE}}}{\partial g_{\phi}(\mathbf{z}_0)} \cdot \frac{\partial g_{\phi}(\mathbf{z}_0)}{\partial \phi}.
    \label{eq:17}
\end{equation}
The update rule is $\phi^{(t+1)} = \phi^{(t)} - \eta_{\phi}\nabla_{\phi}\mathcal{L}_{\mathrm{CE}}$.

\noindent \textbf{LoRA Backbone Update.} 
For the attention weights, let the effective weight be $W = W_0 + \frac{\alpha}{r}AB$. The gradient with respect to the low-rank matrices $\Phi$ is derived via the chain rule through $W$:
\begin{equation}
    \nabla_{\Phi}\,\mathcal{L}_{\mathrm{CE}} = \frac{\partial \mathcal{L}_{\mathrm{CE}}}{\partial W} \cdot \frac{\partial W}{\partial \Phi}.
    \label{eq:18}
\end{equation}
The parameters are updated as $\Phi^{(t+1)} = \Phi^{(t)} - \eta_{\Phi}\nabla_{\Phi}\mathcal{L}_{\mathrm{CE}}$.

\subsubsection{Multi-Turn Optimization}  
At iteration $t$, we first compute $\mathcal{L}_{\mathrm{InfoNCE}}$ to update $\theta$ (Eq.~\ref{eq:12}), ensuring geometric alignment. We then compute the token-level $\mathcal{L}_{\mathrm{CE}}$. Gradients flow simultaneously through the prefix adapter $\phi$ (Eq.~\ref{eq:17}) and the adapter weights $\Phi$ (Eq.~\ref{eq:18}). This alternating optimization decouples semantic anchoring from expressive generation while keeping the LLM backbone frozen.

\section{Prompt Template}\label{sec: prompt}

\subsection{In-Context Prompt Template}
\begin{figure}[!ht]
  \centering
  \includegraphics[width=\columnwidth]{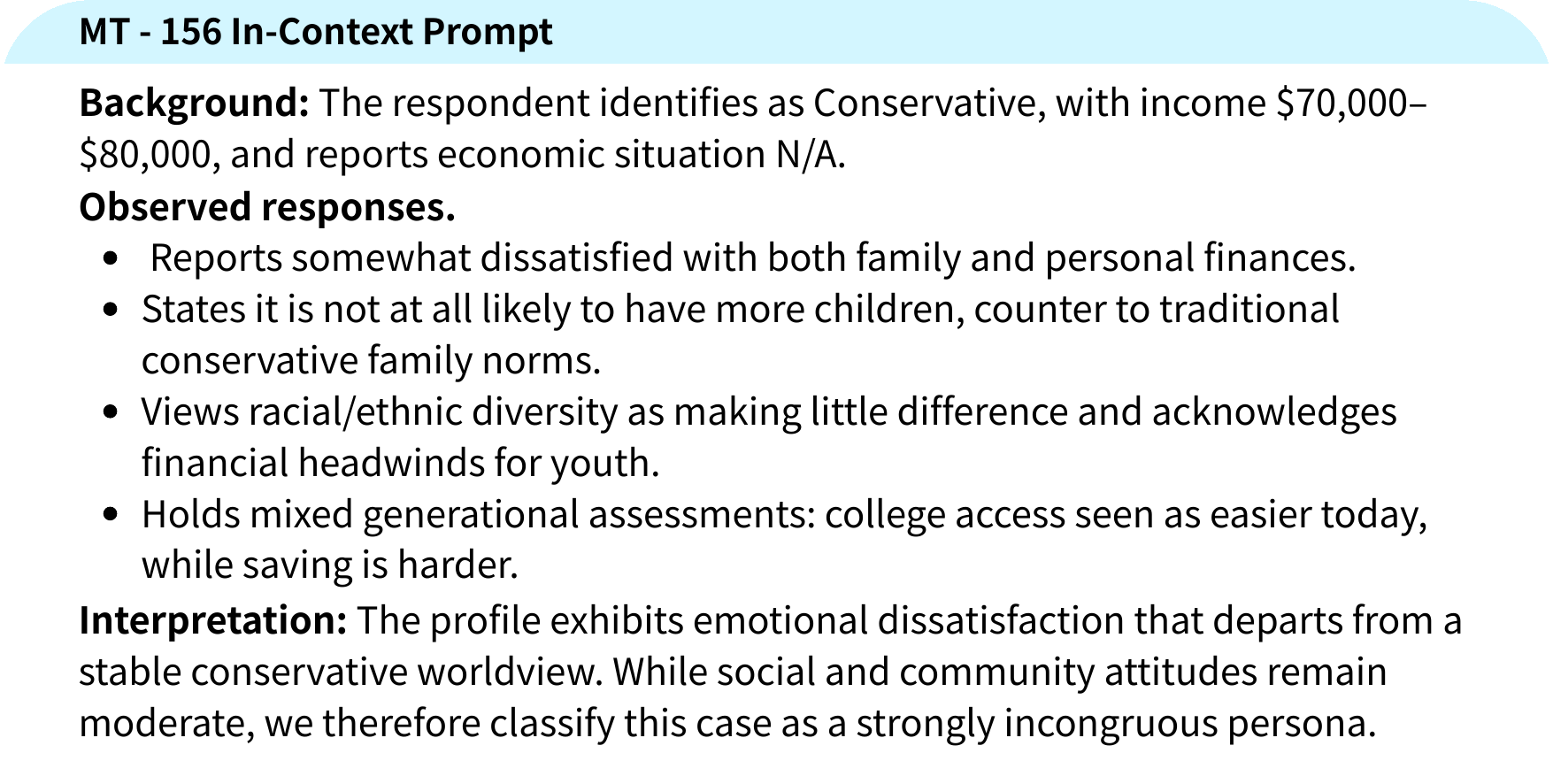}
  \caption{\textbf{In-Context Prompt Template (MT-156).} The structure integrates static backstory attributes with dynamic interaction history.}
  \label{fig:B1}
\end{figure}

Figure~\ref{fig:B1} illustrates the standardized in-context prompt structure. The prompt initializes with the persona backstory $\mathcal{B}$ to establish foundational identity constraints, followed by a sequence of Multiple-Choice Questions (MCQs) that exemplify the persona's historical stances.

\begin{figure}[!ht]
  \centering
  \includegraphics[width=\columnwidth]{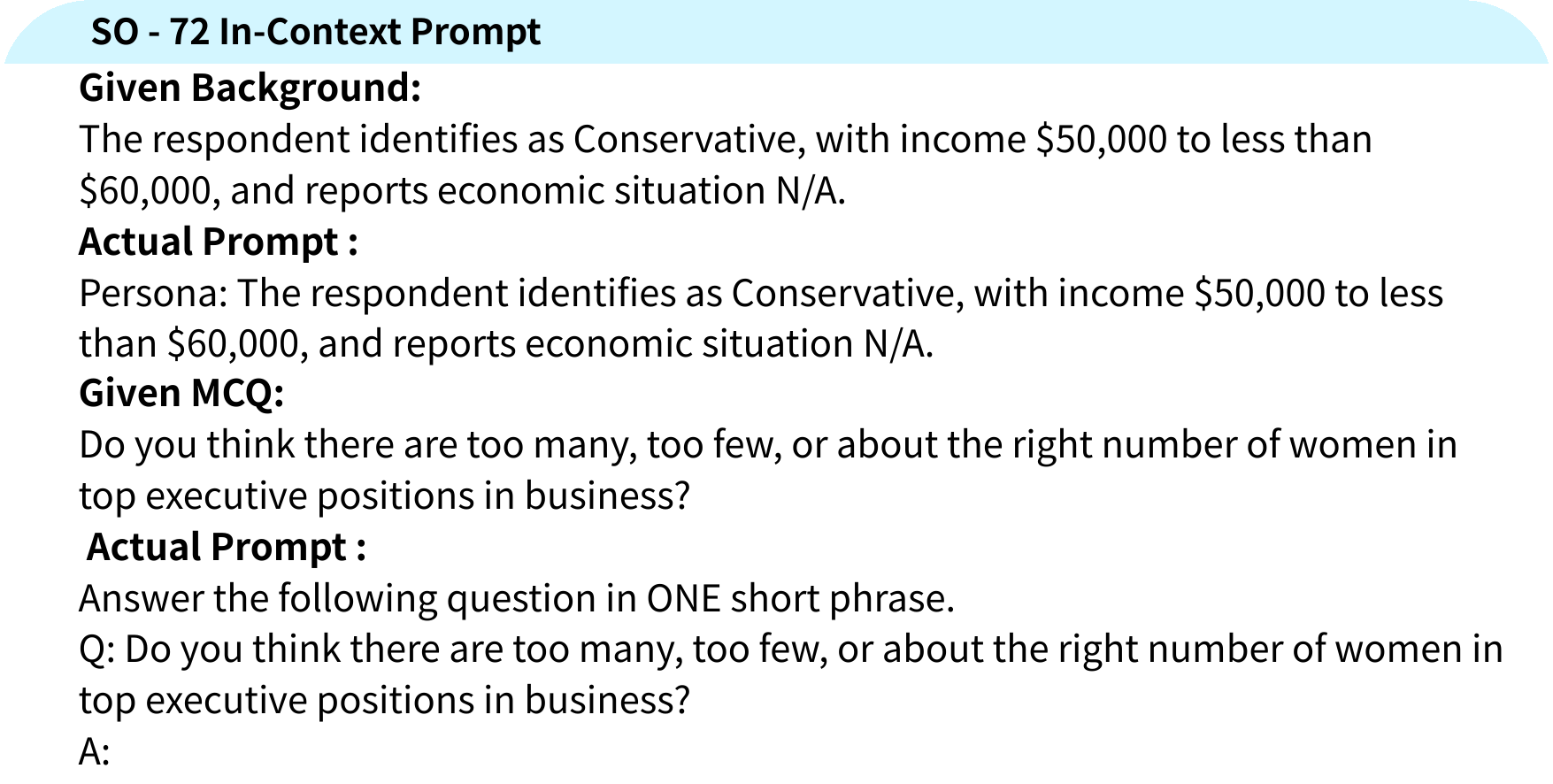}
  \caption{\textbf{Prompt Instantiation Example (SO-72).} A concrete example showing how the persona background and MCQ are formatted into a unified input.}
  \label{fig:Example_prompt}
\end{figure}

\paragraph{Instantiation Example (SO-72).}
To demonstrate the concrete application of our design, Figure~\ref{fig:Example_prompt} presents an instantiation using the \textsc{Social} (SO-72) dataset. The prompt explicitly conditions generation on both the persona background and the specific query constraints. This format enforces ideological alignment, mitigating the model's tendency to revert to generic or persona-agnostic outputs.
\noindent

\subsection{Progressive In-Context Prompt Template}
\begin{figure}[!ht]
  \centering
  \includegraphics[width=\columnwidth]{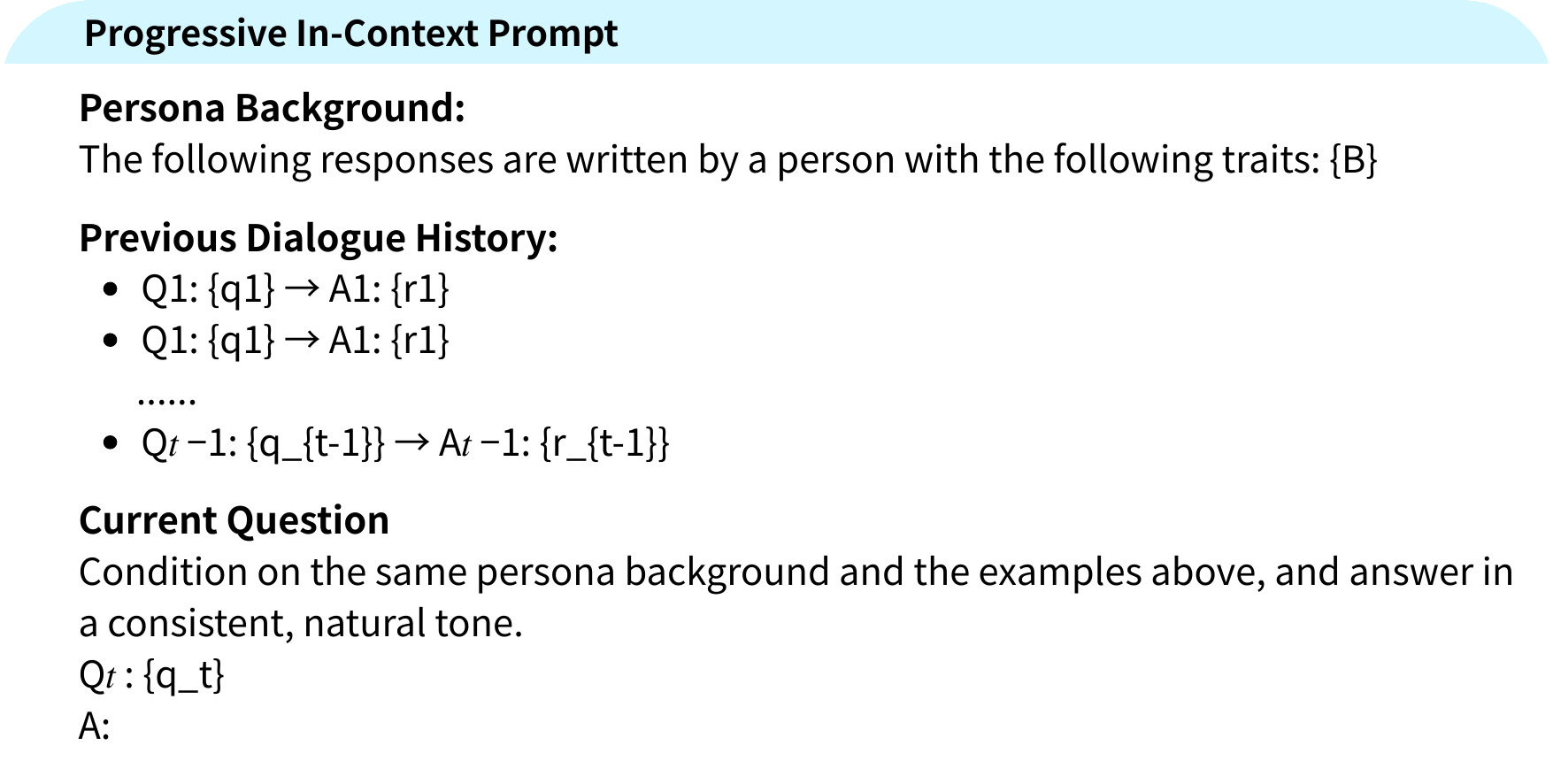}
  \caption{\textbf{Progressive In-Context Template.} The prompt grows dynamically to include history, reinforcing persona consistency over time.}
  \label{fig:B2}
\end{figure}

Figure~\ref{fig:B2} depicts the Progressive In-Context Prompt used during the refinement phase. At training turn $t$, the context window concatenates the persona background $\mathcal{B}$, the cumulative interaction history $\{(q_i, r_i)\}_{i=1}^{t-1}$, and the current query $q_t$. This accumulation strategy encourages the model to internalize stance consistency and reproduce persona-specific patterns across the dialogue trajectory.

\subsection{PQA Inference Template}
\begin{figure}[!ht]
  \centering
  \includegraphics[width=\columnwidth]{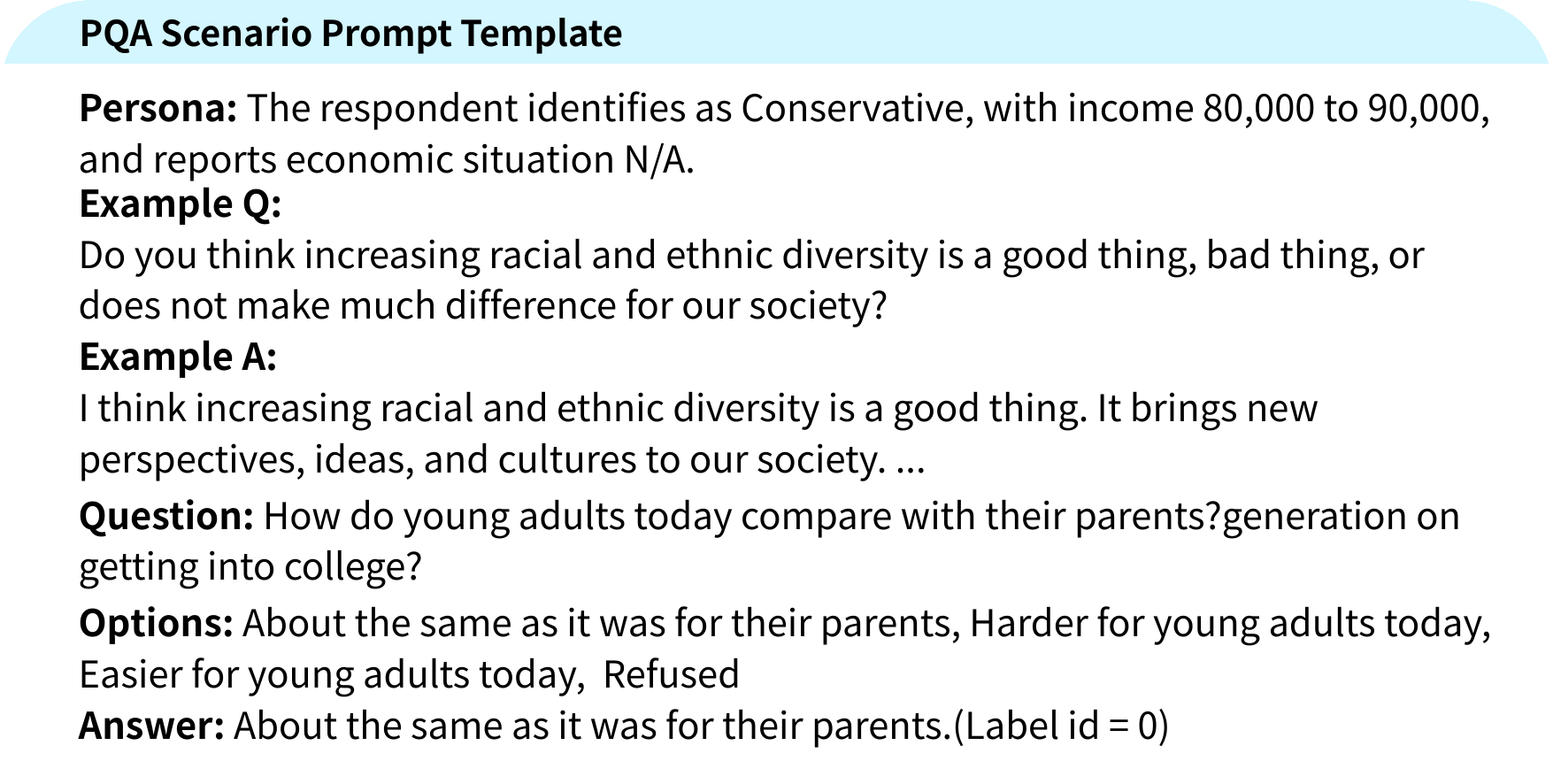}
  \caption{\textbf{PQA Inference Template.} Used for evaluating predictive accuracy on multiple-choice questions.}
  \label{fig:B3}
\end{figure}

Figure~\ref{fig:B3} outlines the inference template for Persona-Steered Question Answering (PQA). Following the refinement phase, the model, conditioned on the stabilized persona state, receives a novel MCQ and predicts the response distribution based on the aligned latent representation.

\subsection{PNI Inference Template}
\begin{figure}[ht]
  \centering
  \includegraphics[width=\columnwidth]{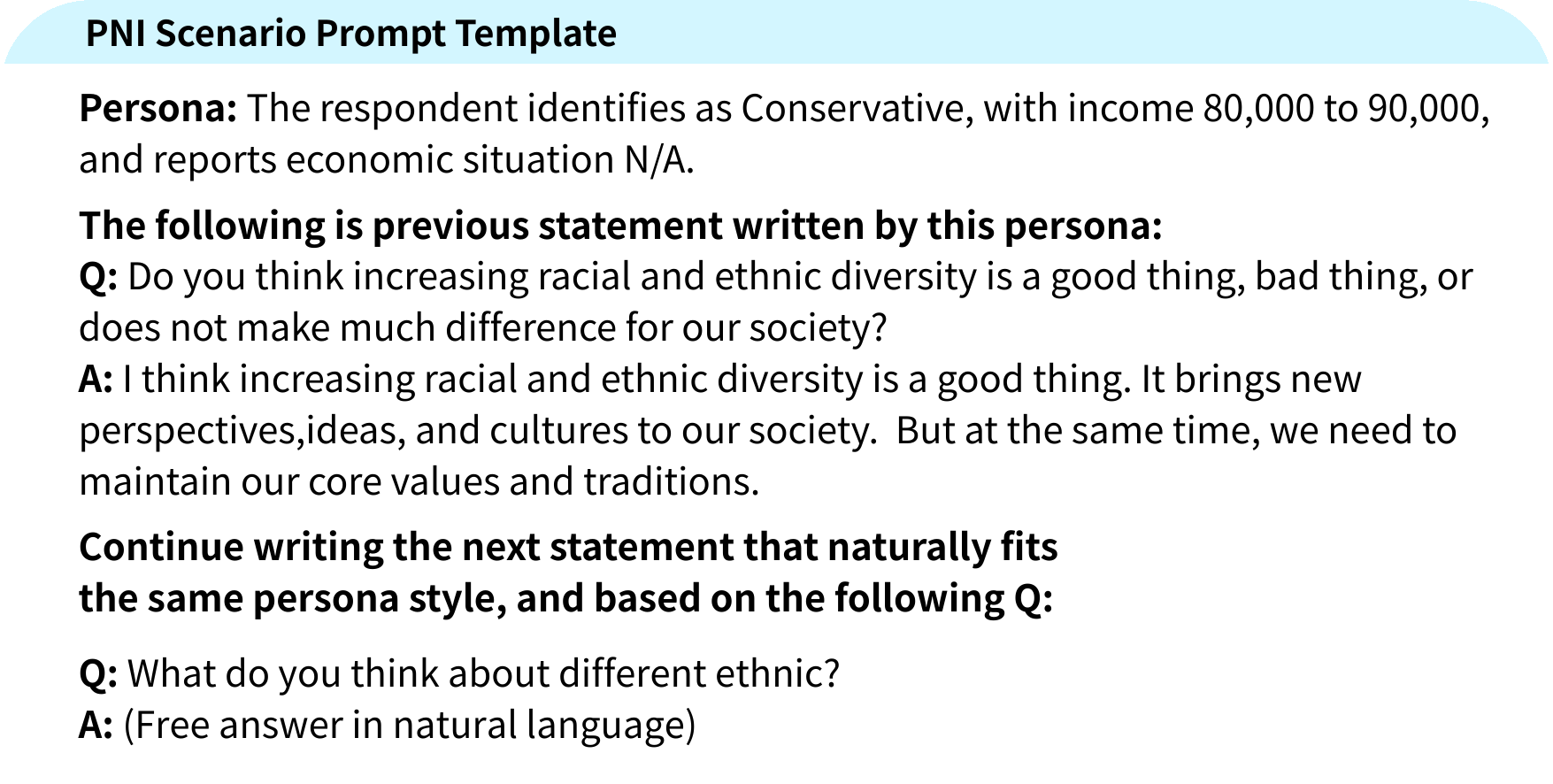}
  \caption{\textbf{PNI Inference Template.} Used for generating and evaluating open-ended narratives.}
  \label{fig:B4}
\end{figure}

Figure~\ref{fig:B4} presents the template for Persona-Steered Narrative Inference (PNI). Unlike the constrained PQA setting, PNI requires open-ended generation. The prompt solicits a free-text response based on the refined persona, which is subsequently evaluated for semantic fidelity and stylistic consistency.

\section{Reproducibility Details}\label{sec: data}

\subsection{Hyperparameter Configuration}\label{subsec:hyperparameters}
We implement QuirkyMind using PyTorch and the Hugging Face Transformers library. Table~\ref{tab:hyperparameters} details the complete configuration. Our training pipeline operates in two distinct phases with differential optimization strategies:

\paragraph{\textbf{Stage 1 (Discriminative Alignment).}} We utilize a higher learning rate ($\eta_\theta = 3 \times 10^{-5}$) to rapidly anchor the persona state in the sentence embedding space via InfoNCE.

\paragraph{\textbf{Stage 2 (Generative Refinement).}} We lower the learning rate ($\eta_\phi = 1 \times 10^{-5}$) to fine-tune the soft prompt adapter and LoRA modules, minimizing the risk of catastrophic forgetting in the backbone.
The multi-objective loss in Stage 2 balances five components. We determine the optimal weighting via grid search on the validation set, setting $\lambda_{\text{KL}} = 0.05$, $\lambda_{\text{Contrast}} = 0.1$, $\lambda_{\text{Anchor}} = 0.1$, and $\lambda_{\text{SoftPrompt}} = 5 \times 10^{-4}$ to manage the trade-off between generation quality and persona consistency.

\begin{table}[t]
\centering
\resizebox{\columnwidth}{!}{%
\begin{tabular}{lr | lr}
\toprule
\multicolumn{2}{c|}{\textbf{Architecture \& Optimization}} & \multicolumn{2}{c}{\textbf{Loss Weights \& Inference}} \\
\midrule
\rowcolor{gray!10}\multicolumn{2}{l|}{\textit{Model Architecture}} & \multicolumn{2}{l}{\textit{Stage 2 Loss ($\lambda$)}} \\
Soft Prompt Tokens ($S$) & 32 & Cross-Entropy (Gen) & 1.0 \\
Soft Prompt Scale ($\alpha$) & 0.5 & Label Smoothing & 0.1 \\
LoRA Rank ($r$) & 4 & KL Divergence & 0.05 \\
LoRA Alpha & 8 & Contrastive (InfoNCE) & 0.1 \\
LoRA Dropout & 0.1 & Anchor Regularization & 0.1 \\
& & Soft Prompt Reg. & 5e-4 \\
\midrule
\rowcolor{gray!10}\multicolumn{2}{l|}{\textit{Optimization (AdamW)}} & \multicolumn{2}{l}{\textit{Inference (Generation)}} \\
Batch Size & 8 & Temperature & 0.7 \\
Weight Decay & 1e-3 & Top-$p$ Sampling & 0.9 \\
LR Stage 1 ($\eta_\theta$) & 3e-5 & Repetition Penalty & 1.1 \\
LR Stage 2 ($\eta_\phi$) & 1e-5 & Max New Tokens & 16 \\
Epochs (Stage 1) & 10 & & \\
Epochs (Stage 2) & 5 & & \\
\bottomrule
\end{tabular}%
}
\caption{\textbf{Hyperparameter Configuration.} We organize the settings into a compact two-pane layout. The left pane details model structure and optimization schedules; the right pane details the multi-objective loss balancing and inference decoding parameters.}
\label{tab:hyperparameters}
\end{table}

\subsection{Evaluation Metrics Details}\label{subsec:evaluation}
For the PNI task, we report three metrics to quantify the trade-off between semantic adherence and lexical variation.

\paragraph{Semantic Faithfulness (BLEU-1).}
To measure how accurately the generated text reflects the ground-truth stance, we calculate sentence-level BLEU-1 with smoothing \cite{papineni-2002-bleu,chen-2014-smoothing}. This metric assesses unigram overlap while penalizing brevity, ensuring the model generates substantive content rather than short, generic acknowledgments.

\paragraph{Lexical Diversity (Distinct-$n$).}
We evaluate the stylistic richness of the generated narratives using Distinct-$1$ (unigrams) and Distinct-$2$ (bigrams) \cite{li-2016-diversity}. These metrics calculate the ratio of unique $n$-grams to the total number of generated tokens:
\begin{equation}
    \text{Distinct-}n = \frac{|\mathcal{U}_n|}{\sum_{i=1}^{N} |h_i|},
\end{equation}
where $\mathcal{U}_n$ represents the set of unique $n$-grams across the entire corpus, and $|h_i|$ denotes the token count of hypothesis $i$. Higher values indicate greater vocabulary usage and less repetitive generation. To ensure comparability, we normalize all generated responses to a fixed maximum length before calculation.

\subsection{Data Pre-processing}  
All experiments read a preprocessed CSV (the repository release will include these CSVs and a short README documenting provenance and mapping). The training script expects each CSV row to already contain the canonical columns used by the pipeline: \texttt{backstory} (long-term persona context), \texttt{MCQ} (question/prompt), \texttt{options} (a list or JSON-encoded list of candidate answers), \texttt{label} (0-based integer index into \texttt{options}), and \texttt{persona\_id} when available. At runtime, the script simply reads the CSV, parses \texttt{options} if it is a string (via \texttt{ast.literal\_eval}), and optionally restricts the dataset to the first \texttt{max\_personas} unique \texttt{persona\_id}s; no additional data-cleaning module is required.

\subsection{Github Repository}
Our anonymized code is available at \url{https://github.com/Chu1004/QuirkyMind}.